\PassOptionsToPackage{numbers}{natbib}
\documentclass{article}

\usepackage[preprint,nonatbib]{neurips_2024}

\usepackage[utf8]{inputenc}      
\usepackage[T1]{fontenc}         
\usepackage{hyperref}            
\usepackage{url}                 
\usepackage{booktabs}            
\usepackage{amsfonts}            
\usepackage{nicefrac}            
\usepackage{microtype}           
\usepackage[dvipsnames]{xcolor}  
\usepackage[numbers]{natbib}
\usepackage{array} 
\usepackage{graphicx}
\usepackage{multirow,multicol,makecell}
\usepackage{pgfplots}
\pgfplotsset{compat=1.18}
\usepackage{pgfplotstable}

\usepackage{siunitx}
\usepackage{datatool}
\DTLsetseparator{=}

\usepackage{graphicx}
\usepackage{float}
\usepackage{subcaption}
\usepackage[rightcaption]{sidecap}

\usepackage[printonlyused]{acronym}

\usepackage{circuitikz,tikz}
\usetikzlibrary{positioning,calc,arrows.meta,math,shapes.misc,plotmarks}

\usepackage{circuitikz,tikz}
\usetikzlibrary{positioning,calc,arrows.meta,math,shapes.misc,plotmarks}

\usepackage{url}
\usepackage{bookmark}

\usepackage[english,noabbrev,nameinlink,capitalise]{cleveref}

\crefname{subfigure}{Figure}{Figures}
\crefname{figure}{Figure}{Figures}
\crefname{table}{Table}{Tables}

\acrodef{VLA}{vision-language-action}
\acrodef{VLM}{vision–language model}
\acrodef{SKU}{stock keeping unit}
\acrodef{MoE}{Mixture-of-Experts}
\acrodef{DP}{Diffusion Policy}
\acrodef{TCP}{Tool Center Point}
\acrodef{PRO}{Process-Reward Operator}

\DeclareSIUnit{\pp}{\%pt}

\title{Cortex 2.0: Grounding World Models in Real-World Industrial Deployment}
\author{%
  \parbox{\textwidth}{\centering
    Adriana Aida \quad Walida Amer \quad Katarina Bankovic \quad
    Dhruv Behl \quad Fabian Busch \\[0.3em]
    Annie Bhalla \quad Minh Duong \quad Florian Gienger \quad
    Rohan Godse \quad Denis Grachev \\[0.3em]
    Ralf Gulde \quad Elisa Hagensieker \quad Junpeng Hu \quad
    Shivam Joshi \quad Tobias Knoblauch \\[0.3em]
    Likith Kumar \quad Damien LaRocque \quad Keerthana Lokesh \quad
    Omar Moured \quad Khiem Nguyen \\[0.3em]
    Christian Preyss \quad Ranjith Sriganesan \quad Vikram Singh \quad
    Carsten Sponner \quad Anh Tong \\[0.3em]
    Dominik Tuscher \quad Marc Tuscher \quad 
    Pavan Upputuri \\[0.5em]
    Sereact GmbH \\[0.3em]
    {\tiny Authors listed in alphabetical order.}
  }%
}

\begin{document}
\vspace*{-2cm}

\maketitle

\begin{figure}[ht]
  \centering
  \includegraphics[width=\textwidth]{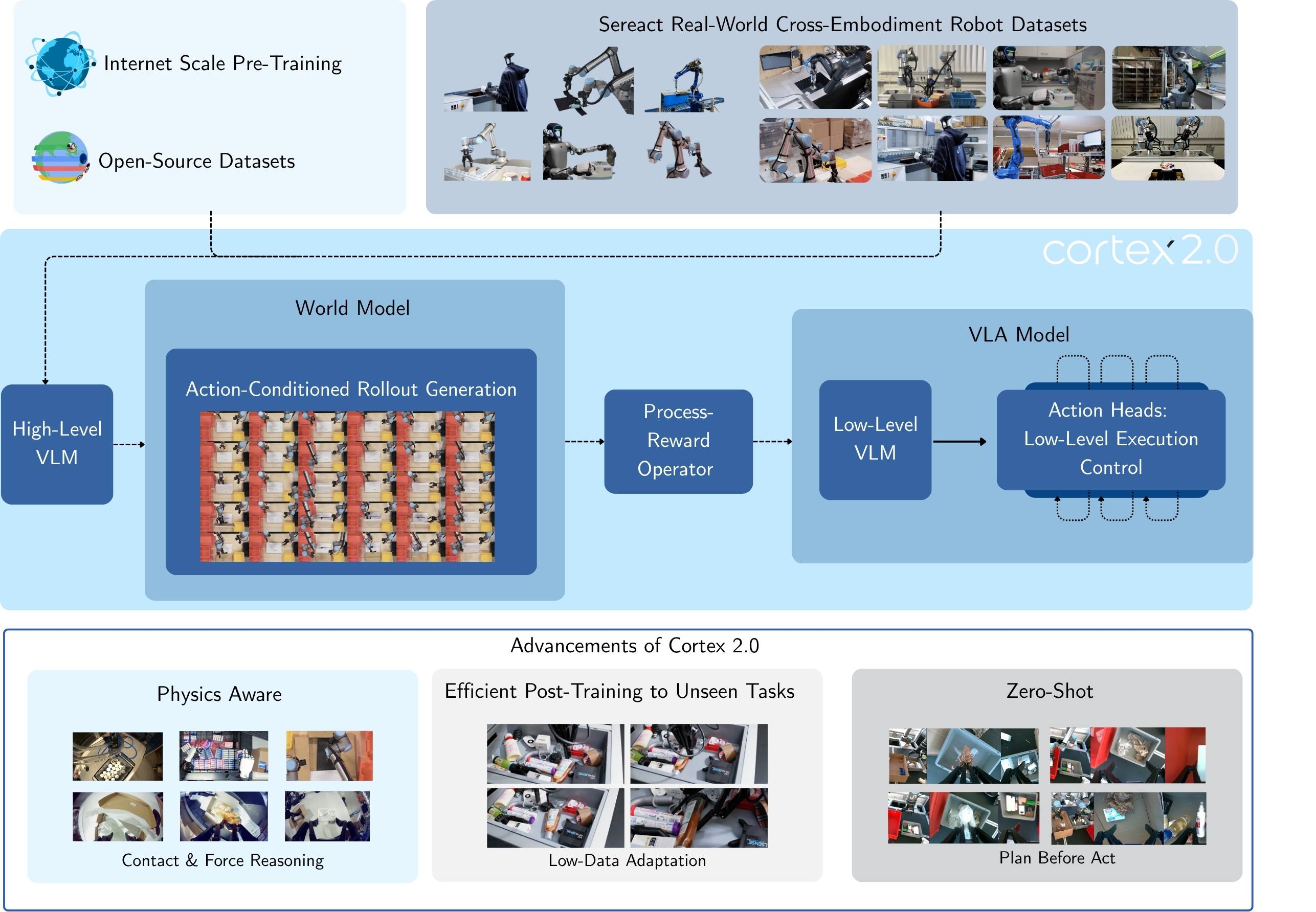}
  \caption{Overview of Cortex 2.0}
\end{figure}



\begin{abstract}
Industrial robotic manipulation demands reliable long-horizon execution across embodiments, tasks, and changing object distributions. While Vision-Language-Action models have demonstrated strong generalization, they remain fundamentally reactive. By optimizing the next action given the current observation without evaluating potential futures, they are brittle to the compounding failure modes of long-horizon tasks. Cortex 2.0 shifts from reactive control to plan-and-act by generating candidate future trajectories in visual latent space, scoring them for expected success and efficiency, then committing only to the highest-scoring candidate. We evaluate Cortex 2.0 on a single-arm and dual-arm manipulation platform across four tasks of increasing complexity: pick and place items, item and trash sorting, screw sorting, and shoebox unpacking. Cortex 2.0 consistently outperforms state-of-the-art Vision-Language-Action baselines, achieving the best results across all tasks. The system remains reliable in unstructured environments characterized by heavy clutter, frequent occlusions, and contact-rich manipulation, where reactive policies fail. These results demonstrate that our world-model-based planning can operate reliably in complex industrial environments. 
\end{abstract}

\section{Introduction}
\label{sec:intro}
Reliable robotic manipulation at industrial scale requires more than generalization. Actions are irreversible, failures compound over long horizons, and a single unrecovered error can disrupt an entire production workflow. While recent Vision--Language--Action (VLA) models~\cite{pi0, pi05, rt1} have demonstrated strong generalization across tasks and embodiments, they remain reactive by design: each action is selected conditioned on the current observation, without explicit reasoning about future outcomes.

Cortex~2.0 extends our preceding VLA model, Cortex, with a world-model-based planning module. At each decision step, a world model generates a set of candidate future trajectories in visual latent space. The Process-Reward Operator (PRO), our dense reward module, scores each candidate for task progress, risk likelihood, and completion likelihood, and the policy commits to the highest-scoring trajectory. Before  action, the system evaluates potential futures rather than committing to the first available action. 

\paragraph{Background and related progress.}
Large-scale robot demonstration data and cross-embodiment corpora have driven great progress in manipulation policies~\cite{rt2, palme, OXE2024, Bridge2022}. Sereact's operating fleet has accumulated large-scale manipulation data across warehouse deployments, providing training data that reflects real industrial conditions including edge cases and failure modes that are difficult to reproduce in controlled data collection. Alongside data scaling, world models have emerged as a promising approach: UniSim~\cite{unisim} and Cosmos~\cite{cosmos} have shown that models trained on internet-scale video acquire broad physical priors transferable to robotic settings. Cortex~2.0 builds on this direction, grounding world model training in deployment data collected continuously from live operations.

\paragraph{Industrial setting.}
Warehouse manipulation involves frequent occlusions from totes and packaging, reflective and translucent surfaces that challenge RGB-based perception, and rapid object distribution shift across shifts and sites. 
In tasks such as returns handling, failure modes including gradual slip, jams, and collisions emerge only after several steps. World-model-based planning addresses this: by scoring candidate futures before execution, the system can identify and avoid problematic branches prior to commitment.

Our main contributions are:
\begin{enumerate}
    \item \textbf{World-model augmented VLA:} we integrate a visual-latent world model into Cortex, enabling k-step lookahead planning. 
    \item \textbf{PRO scoring module:} we introduce a multi-criteria scoring function that evaluates candidate rollouts for task progress, completion likelihood, and risk likelihood. It thereby derives an advantage signal that conditions the action heads. 
    \item \textbf{Cross-embodiment planning:} because planning operates in visual space, the same planning loop transfers across single-arm, dual-arm, and humanoid embodiments. 
    \item \textbf{Benchmark evaluation:} we benchmark Cortex 2.0 on four real-world tasks of increasing complexity, achieving highest success rates across all tasks with zero human interventions. 
\end{enumerate}

\section{Related Works}
\subsection{Vision--Language--Action Models}

Since RT-1~\cite{rt1} and RT-2~\cite{rt2}, large sequence models have served as generalist robot policies by mapping camera observations and language prompts to actions. RT-2 demonstrated that pretraining on internet-scale vision--language data transfers semantic knowledge to robotic control, establishing the VLA template that subsequent work has built upon. Generalist policies such as Octo~\cite{octo} and OpenVLA~\cite{openvla2024} scaled this template to cross-embodiment pretraining; OpenVLA-OFT~\cite{Kim2025OpenVLA-OFT} further showed that parallel decoding and action chunking substantially improve inference speed without sacrificing task performance. PaLM-E~\cite{palme} further showed that embodied multimodal language models can ground high-level reasoning in physical scenes.

Recent systems increasingly adopt hierarchical designs that predict mid-level subtask tokens before executing fine-grained control~\cite{pi0,pi05, molmoact2025, f1_vla_2025}. $\pi_0$~\cite{pi0} instantiated a flow-matching action expert on top of a VLM backbone and demonstrated strong dexterous manipulation across diverse platforms. $\pi_{0.5}$~\cite{pi05} extended this by co-training discrete action tokens with web and language data, improving generalization to unseen environments. FAST~\cite{pertsch2025fast} introduced efficient action tokenization for high-frequency autoregressive control, and GR00T N1~\cite{groot2025} demonstrated hierarchical dual-system VLA architectures for humanoid platforms.
The preceding Cortex system \cite{sereact2025cortex} introduced a three-level VLA design for industrial settings, combining the Sereact Lens VLM for subtask prediction and pixel-level grounding with flow-matching action heads, and demonstrated strong performance on warehouse pick-and-place and returns handling tasks. Cortex~2.0 extends this by augmenting the reactive policy with world-model-based future predictions, moving from pattern-matched responses to informed plan selection.

\subsection{Flow Matching for Robot Control}

Diffusion policies~\cite{diffusionpolicy, ho2020ddpm} model action generation as a denoising process, generating expressive multimodal policies for manipulation. Flow matching~\cite{lipman2022flow, albergo2022building, albergo2023stochastic} improves on this by learning straight-line interpolations between noise and data, reducing the number of inference steps required. $\pi_0$~\cite{pi0} was the first large-scale VLA to adopt flow matching for action generation, with $\pi_{0.5}$~\cite{pi05} and RDT-2~\cite{liu2026rdt2} further validating its advantages in latency and trajectory quality in diverse bimanual tasks.

\subsection{World Models for Robotics}

World models are predictive models of environment dynamics with a long history in model-based RL. Early work~\cite{ha2018world, finn2017deep} formalized world models for policy learning; Dreamer~\cite{hafner2020dreamer, hafner2023dreamerv3} established that latent imagination can match model-free approaches on visual control tasks, though such approaches use the world model as a training-time rollout generator, which risks compounding model errors.

At internet scale, UniSim~\cite{unisim} and Cosmos~\cite{cosmos} demonstrated that world models pretrained on large-scale video acquire broad physical priors transferable to robotic settings. Several concurrent works have explored using such models at inference time: IRASim~\cite{zhu2025irasim} and GPC~\cite{gpc2025} showed that scoring candidate rollouts before execution improves task success over reactive policies, while GR-2~\cite{gr2_2024} and V-JEPA~2~\cite{assran2025vjepa2} validated that joint pretraining on internet video and robot data supports strong physical reasoning with limited robot-specific supervision. Li et al.~\cite{li2025robotic} further demonstrated this direction on deployment data. Cortex~2.0 builds on these findings by grounding world model training in continuously collected operational data and scoring imagined rollouts via PRO before any action is executed.

Cortex~2.0 follows this direction: the world model is pretrained on internet-scale video and fine-tuned on deployment recordings at $30\,\text{Hz}$. Central to this design is that imagined futures remain grounded in the same representational space as real observations, so that PRO's scoring function, learned on real executed trajectories, transfers directly to imagined futures.

\subsection{Force Feedback and Multimodal Sensing}

RGB-only policies are inherently limited in scenarios involving contact, deformation, and occlusion. Force and tactile sensing provide complementary signals that are invisible to cameras: contact forces reveal grasp stability, torque profiles encode object compliance, and vacuum pressure margins indicate suction reliability. Early work demonstrated that force–torque signals enable more robust grasping under uncertainty~\cite{calandra2018more}, and subsequent systems have shown that fusing vision and touch improves performance on contact-rich tasks~\cite{Lee2020}.

In Cortex~2.0, force feedback is an optional input used only when the robot supports it. When available, the robot state $r_t$ and force feedback $f_t$ are added to the multimodal observation $o_t$ (Eq.~\ref{eq:obs}) alongside RGB and task instruction embeddings. This design allows Cortex~2.0 to deploy across platforms with heterogeneous sensing configurations: contact-rich tasks such as screw sorting or kitting benefit from force-aware observations. Suction-based tasks benefit from vacuum pressure signals that reveal grasp reliability not visible to cameras. 

\subsection{Datasets for Robot Learning}
\label{sec:rw-dataset}

Scale and diversity are crucial for generalization and cross-embodiment transfer. Bridge and BridgeData V2 study cross-domain mixtures for manipulation transfer \cite{Bridge2022,Bridgev22023}. Open X-Embodiment unifies demonstrations across many labs and robot embodiments to enable training of generalist policies \cite{OXE2024}. DROID and Agibot World further enlarge the distribution toward in-the-wild and large-scale manipulation \cite{khazatsky2024droid,bu2025agibot_arxiv}. Domain-specific corpora such as Stanford Kuka and Berkeley cable routing provide long-horizon and contact-rich tasks \cite{Lee2020,Luo2024}. Despite breadth, few public datasets capture industrial warehouse conditions; many production datasets remain proprietary.

\paragraph{Positioning.}
Our approach builds on the VLA model with flow-matching action heads. We augment our previous reactive policy with a visual-latent world model that generates $k$ candidate futures at inference time, scored by our Process-Reward Operator (PRO) for task progress, risk, and completion likelihood before any action is executed. In addition to public datasets, training incorporates a proprietary dataset of deployment recordings collected continuously across our deployments.
\section{Methodology}
\label{sec:methodology}
\label{sec:cortex-framework}
\acresetall

\subsection{Overview}

\begin{figure}[ht]
  \centering
  \includegraphics[width=\textwidth]{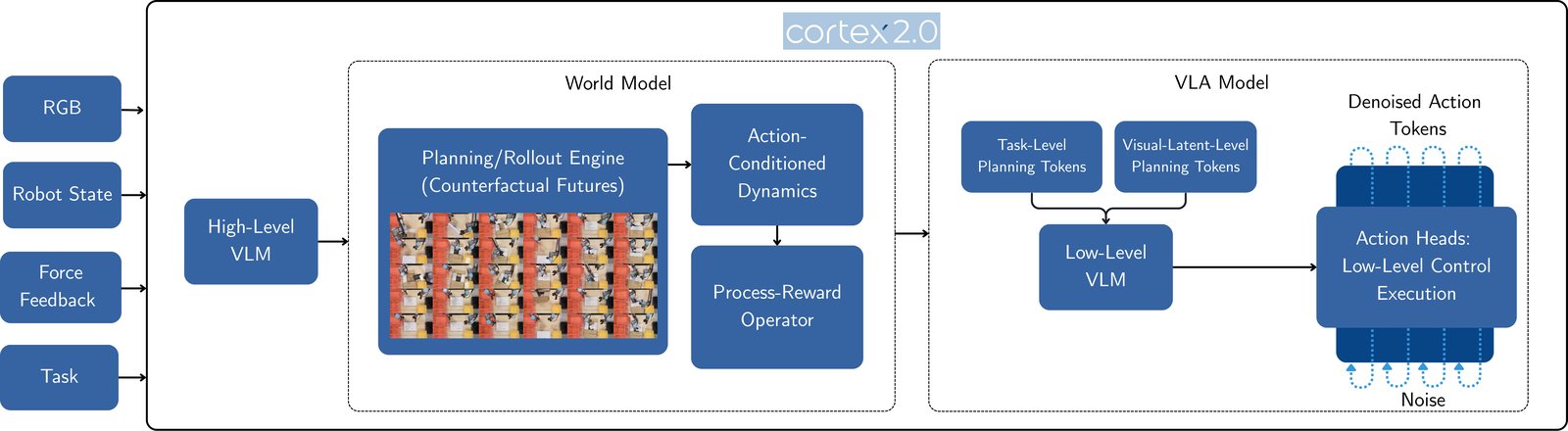}
  \caption{Cortex 2.0 Architecture}
  \label{fig:arch}
\end{figure}

Cortex 2.0 is Sereact's general-purpose vision--language--action (VLA) model, validated in industrial applications such as pick and place, returns handling, and kitting.
It unifies perception, planning, reasoning, and control in a four-level hierarchical design: a high-level VLM observes and encodes the scene; a world model generates candidate futures; PRO evaluates and ranks them; and flow-based action heads commit to the highest-scoring trajectory (Figure~\ref{fig:arch}).

We consider a partially observed control problem with discrete time $t$ = 1,...,$T$. At each step, the robot receives multimodal observations 

\begin{equation}
    o_t = ({I_t^{rgb}, r_t, f_t, l_t}),
    \label{eq:obs}
\end{equation}

where $I_t^{rgb}$ is the RGB  image from wrist cameras, $r_t$ is the robot state, $f_t$ the force feedback of the end-effector, and $l_t$ the embedding of the task instruction. 
The observation is encoded into a visual latent $z_t = f_{\text{enc}}(o_t)$, where $f_{\text{enc}}$ is the VLM visual encoder.

A high-level VLM produces structured task context $s_t$ from $z_t$. The world model generates $k$ future candidate trajectories over planning horizon $H_{\text{wm}}$, which the PRO module scores and ranks. The selected plan $\tau^*$ and its binarized advantage indicator $I_t \in \{0,1\}$ condition the policy 
$\pi_\theta$, which produces an action chunk over execution horizon $H_{\text{act}}$:

\begin{equation}
    \pi_{\theta}\!\left(a_{t:t+H_{\text{act}}-1} \mid 
    z_t,\, s_t,\, \tau^*,\, I_t\right).
    \label{eq:policy}
\end{equation}

Formally, at each decision step $t$ the system solves:
\begin{equation}
    \tau^* = \arg\max_{\tau_j \in \{\tau_1, \ldots, \tau_k\}} S_j(z_t, s_t),
\end{equation}

where $S_j \equiv S(\tau_j)$ is the PRO score of rollout $\tau_j$. 
The advantage of the selected rollout relative to the average candidate score is defined as:
\begin{equation}
    \Delta^* = S(\tau^*) - \frac{1}{k}\sum_{j=1}^{k} S(\tau_j),
    \label{eq:advantage}
\end{equation}
and binarized via a task-dependent threshold $\epsilon(s_t)$ into an indicator $I_t = \mathbf{1}[\Delta^* > \epsilon(s_t)]$, which conditions the VLA policy.

Given demonstration trajectories \(\mathcal{D} = \{(o_{1:T}, a_{1:T})\}\), training jointly optimizes:
\begin{equation}
    \mathcal{L}_{\text{total}}(\theta)
    =
    \mathcal{L}_{\text{FM}}(\theta)
    + \lambda_{\text{wm}} \mathcal{L}_{\text{WM}},
    \label{eq:total_loss}
\end{equation}
where $\mathcal{L}_{\text{FM}}$ is the flow-matching action loss and
$\mathcal{L}_{\text{WM}}$ the world model loss. PRO is pretrained separately on industrial deployment data and kept frozen during this stage. 

Cortex 2.0 is trained on open-source multimodal datasets, Sereact's teleoperation and deployment data, and synthetic data, enabling the model to generalize across embodiments and task families under real-world conditions.

\subsection{Cortex 2.0 Architecture}

\subsubsection{High-Level VLM}
The high-level VLM encodes the current observation into a structured task context that mediates between perception, planning, and control. Given latent \(z_t\), it produces:
\begin{equation}
s_t = f_{\mathrm{hl-VLM}}(z_t),
\end{equation}
where \(s_t\) is a learned task-conditioned embedding that encodes a subgoal-level decomposition of the task and grounding variables aligning language with scene entities and spatial constraints, including objects, spatial relations, and contact priors. This representation steers the world model toward task-relevant futures and the execution stack toward realizing the selected future as action. 

\subsubsection{PRO: Process-Reward Operator}

PRO is our dense reward model operating over executed trajectories from real deployment data. Cortex 2.0 now lifts PRO into the planning loop: PRO heads operate on the visual latents $z_{t+1:t+H_{\text{wm}}}^{(j)}$ predicted by the world model (Section~\ref{sec:wm}), scoring imagined rollouts before any action is executed.

For each candidate rollout \(j\), PRO consumes the sequence of predicted latents
\begin{equation}
    \hat{\tau}_j = \left(z_{t+1}^{(j)},\, z_{t+2}^{(j)},\, \ldots,\, 
    z_{t+H_{\text{wm}}}^{(j)}\right).
    \label{eq:pro_input}
\end{equation}

A temporal model processes this latent sequence to produce a rollout-level representation \(p_j\) that captures both the predicted final state and the quality of the trajectory leading to it. From \(p_j\), three prediction heads operate.

\paragraph{Progress.}
The progress head estimates how much closer the predicted future brings the system to successful task completion:
\begin{equation}
    \Delta p^{(j)} = V_{\phi}\!\left(z_{t+H_{\text{wm}}}^{(j)}\right) 
    - V_{\phi}(z_t),
    \label{eq:progress}
\end{equation}
where \(V_{\phi}\) is a value function learned over visual latents from executed trajectories with known outcomes.

\paragraph{Risk.}
The risk head predicts the probability of a failure event occurring along the imagined trajectory:
\begin{equation}
    \rho^{(j)} = P_{\phi}\!\left(\text{fail} = 1 \mid \hat{\tau}_j\right),
    \label{eq:risk}
\end{equation}
penalizing rollouts that pass through latent states associated with high-speed contact, compression, edge impacts, or surface scraping, even if the item ultimately reaches the goal.

\paragraph{Termination.}
The termination head predicts the likelihood that the imagined trajectory leads to successful task completion:
\begin{equation}
    d^{(j)} = P_{\phi}\!\left(\text{success} = 1 \mid \hat{\tau}_j\right).
    \label{eq:term}
\end{equation}

\begin{figure}[h]
    \centering
    \includegraphics[width=\linewidth]{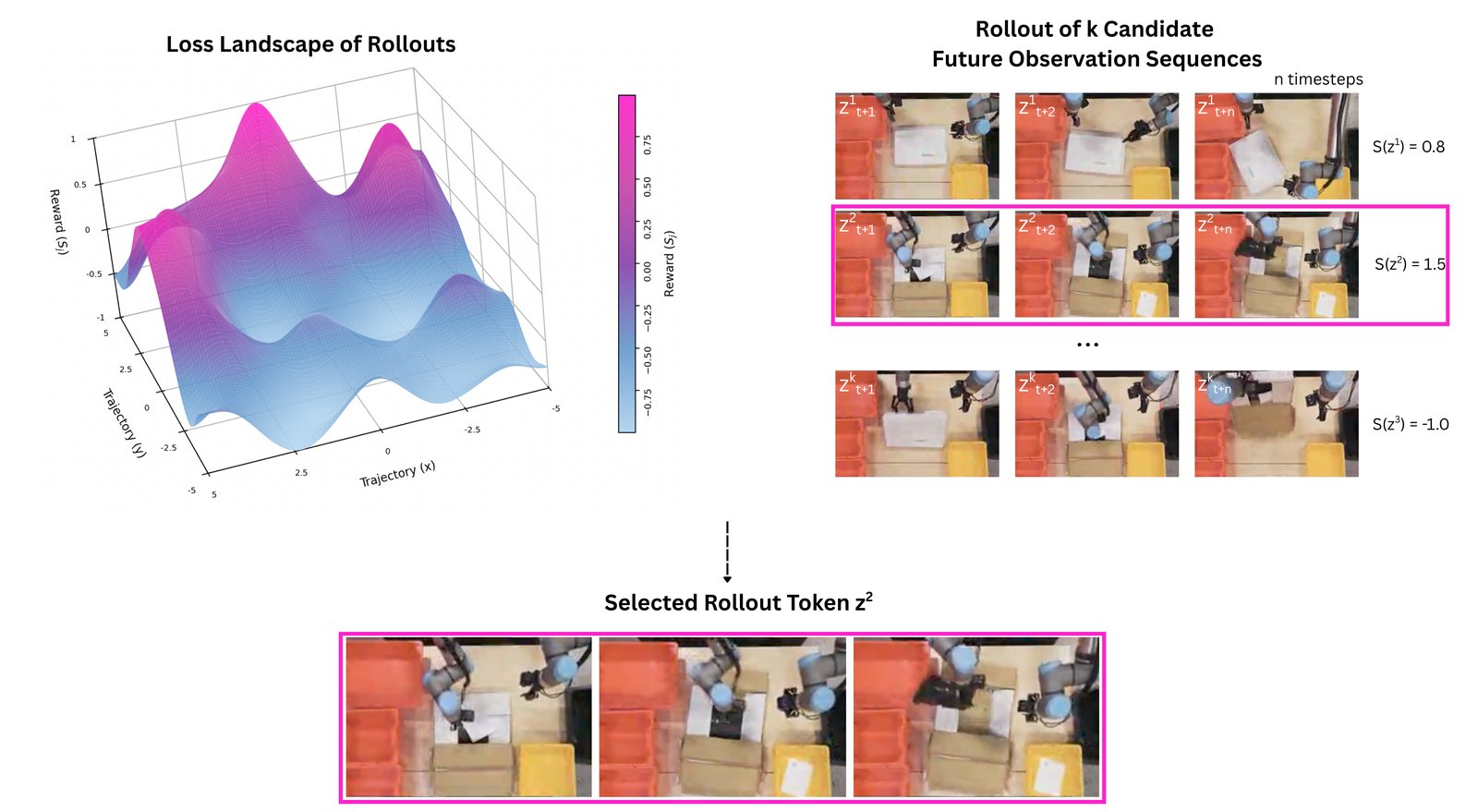}
    \caption{PRO scores $k$ candidate rollouts via the composite score $S_j$ 
    (Eq.~\ref{eq:composite}). The loss landscape shows all candidate trajectories (top); 
    PRO selects the highest-scoring rollout $\tau^*$ (bottom). }
    \label{fig:pro_selection}
\end{figure}

PRO aggregates these three signals into a composite rollout score:
\begin{equation}
    S_j = \Delta p^{(j)} - \lambda\, \rho^{(j)} + \beta\, d^{(j)},
    \label{eq:composite}
\end{equation}
where $\lambda$ controls risk sensitivity and $\beta$ weights completion 
likelihood. Figure~\ref{fig:pro_selection} illustrates the full scoring 
and selection process across $k$ candidate rollouts.

As illustrated in Figure~\ref{fig:pro_selection}, the best rollout is then selected as:
\begin{equation}
    z_*^{\text{token}} = \arg\max_j\; S_j.
\end{equation}

The advantage $\Delta^*$ (Eq.~\ref{eq:advantage}) is binarized via threshold $\epsilon(s_t)$ into indicator $I_t = \mathbf{1}[\Delta^* > \epsilon(s_t)]$, which is passed to the VLA policy. 
At inference, $I_t$ is fixed to $1$, biasing the VLA to generate actions that best realize the selected future.

The PRO heads are trained on real executed trajectories from deployment data, where ground-truth outcomes are available, and applied at inference time to imagined latents from the world model. PRO is pretrained and kept frozen during world model and policy training. This transfer is enabled by the world model producing latents in the same visual latent space as real observations.

\subsubsection{World Model}\label{sec:wm}
The world model $f_{\varphi}$ learns predictive dynamics in visual latent space via flow matching~\cite{lipman2022flow}. Conditioned on the current latent $z_t$ and task context $s_t$, it generates $k$ candidate future latent sequences, each initialized from a distinct noise realization 
$\xi^{(j)} \sim \mathcal{N}(0, I)$, which PRO then scores to select the highest-quality trajectory.

\paragraph{Training.}
For each ground-truth future latent $z_{t+h}$ at step $h \in 
\{1,\ldots,H_{\text{wm}}\}$, conditioned on $z_t$ and $s_t$, we sample 
flow time $\sigma \sim \mathrm{Beta}(\alpha, \beta)$ with $\beta \gg \alpha$, and noise $\xi^{(h)} \sim \mathcal{N}(0, I)$ independently per step, forming the interpolation:
\begin{equation}
    \tilde{z}_\sigma^{(h)} = \sigma\,z_{t+h} + (1 - \sigma)\,\xi^{(h)},
    \qquad v^{(h)} = z_{t+h} - \xi^{(h)}.
    \label{eq:wm_interp}
\end{equation}
Choosing $\beta \gg \alpha$ biases sampling toward higher noise levels, which allows reliable trajectory scoring with fewer ODE integration steps at inference. Since PRO operates on motion and physical plausibility rather than rendering fidelity, coarse latent reconstruction are sufficient to distinguish good from bad trajectories. 
The world model is trained with the flow-matching objective:
\begin{equation}
    \mathcal{L}_{\text{WM}}(\varphi)
    =
    \mathbb{E}_{h,\,\sigma,\,\xi^{(h)}}
    \left\|
    g_\varphi\!\left(\tilde{z}_\sigma^{(h)},\, \sigma,\,
    z_{t},\, s_t\right)
    - v^{(h)}
    \right\|_2^2 .
    \label{eq:wm_loss}
\end{equation}

\begin{figure}[h]
    \centering
    \includegraphics[width=0.85\linewidth]{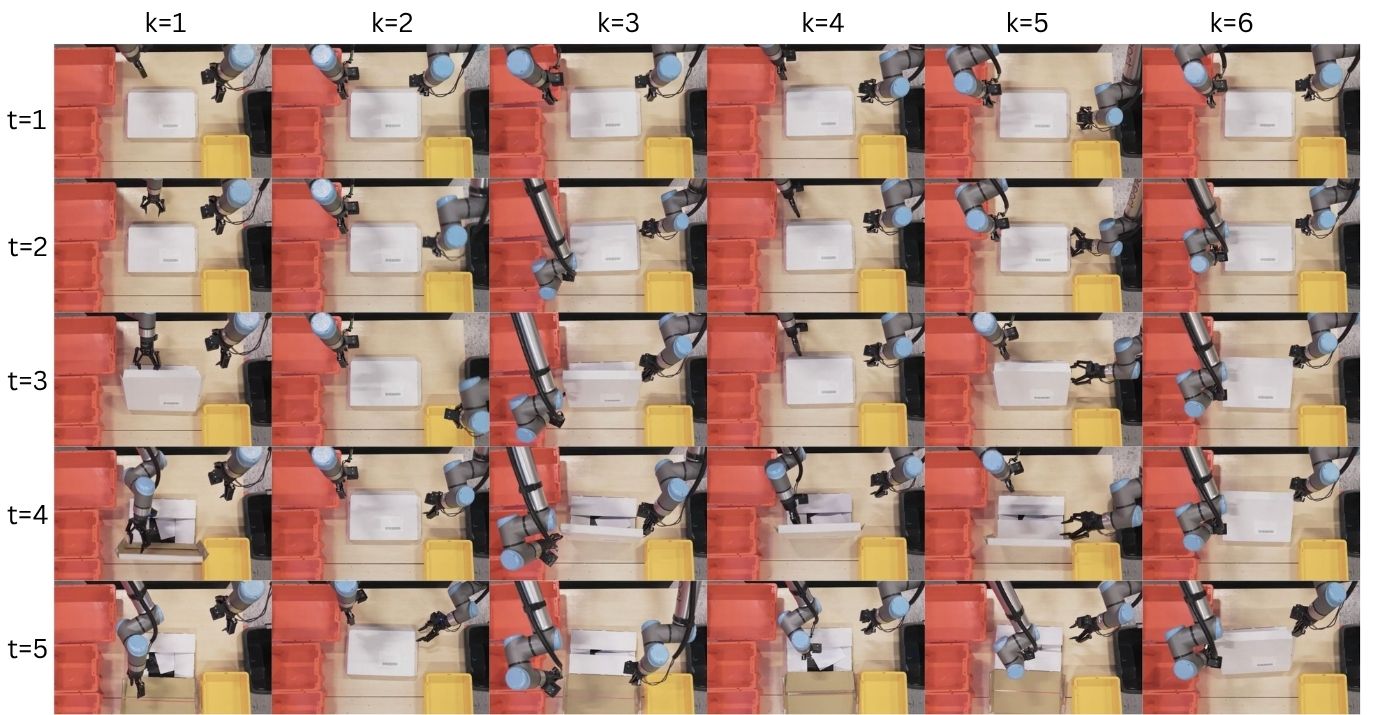}
    \caption{World Model Rollout Generation. Each column of the $6\times5$ grid shows one of $k{=}6$ candidate trajectories rolled out over horizon $H_{\text{wm}}=5$. The rollouts share the same current latent $z_t$ and task context $s_t$, but each is generated from an independent noise realization $\xi^{(j)} \sim \mathcal{N}(0, I)$. PRO scores each rollout and selects the highest-ranked candidate $\tau^*$ for execution.}
    \label{fig:rollouts}
\end{figure}

\paragraph{Inference.}
For each candidate $j = 1,\ldots,k$, future latents are generated. A noise realization $\xi^{(j)} \sim \mathcal{N}(0,I)$ is sampled and the ODE is integrated from $\sigma = 0$ to $\sigma = 1$:
\begin{equation}
    \tilde{z}_{\sigma+\Delta\sigma}^{(j)}
    =
    \tilde{z}_\sigma^{(j)}
    +
    \Delta\sigma \cdot g_\varphi\!\left(
        \tilde{z}_\sigma^{(j)},\, \sigma,\,
        z_t,\, s_t
    \right),
    \label{eq:wm_ode}
\end{equation}
yielding the full candidate sequence $z_{t+1:t+H_{\text{wm}}}^{(j)} = \tilde{z}_{\sigma=1}^{(j)}$, which is passed to PRO.
Figure~\ref{fig:rollouts} illustrates the generation of 30 future candidate sequences.

The world model is pretrained on internet-scale video data and fine-tuned on deployment recordings at $30\,\mathrm{Hz}$. The number of ODE integration steps controls the accuracy of the numerical integration, trading inference speed against trajectory fidelity.

\subsection{VLA Policy}\label{sec:vla}

The VLA policy receives the task context $s_t$, the current visual latent $z_t$, the selected world model rollout $z_*^{\text{token}}$, and the advantage signal $I_t \in \{0,1\}$ from PRO, and generates a continuous action chunk. 
The policy is implemented as a 2B-VLM with a flow-matching action head.

The selected rollout token and advantage indicator are projected into the VLM feature space and concatenated with the task context:
\begin{equation}
    c_t = \left[s_t;z_t;\; W_z\, z_*^{\text{token}};\; W_I\, I_t\right],
    \label{eq:context}
\end{equation}
where $W_z \in \mathbb{R}^{d \times d_z}$ and $W_I \in \mathbb{R}^{d \times 1}$ 
are learned projections. This conditioning is processed through the full depth of the VLM transformer.

\paragraph{Flow-Matching Action Head.}
Given a ground-truth action chunk $x \in \mathbb{R}^{H_{\text{act}} \times C}$~\cite{zhao2023act} and conditioning $c_t$, we sample $\nu \sim \mathrm{Beta}(\alpha, \beta)$ with $\alpha{=}1.0$, $\beta{=}1.5$ (rescaled via $\nu \leftarrow 0.001 + 0.999\,\nu$) and noise $\varepsilon \sim \mathcal{N}(0, I)$. The interpolation and target velocity are:
\begin{equation}
    x_\nu = \nu x + (1-\nu)\varepsilon, \qquad u_\nu = x - \varepsilon.
\end{equation}
The action head predicts velocity $v_\theta(x_\nu, \nu, c_t)$, trained with:
\begin{equation}
    \mathcal{L}_{\mathrm{FM}}(\theta) = 
    \mathbb{E}_{(x,c_t)\sim\mathcal{D}}\;
    \mathbb{E}_{\nu\sim\rho}\;
    \mathbb{E}_{\varepsilon\sim\mathcal{N}(0,I)}
    \left\| v_\theta(x_\nu, \nu, c_t) - u_\nu \right\|_2^2.
\end{equation}
At inference, actions are generated by integrating the ODE forward from noise:
\begin{equation}
    x_{\nu+\Delta\nu} = x_\nu + \Delta\nu \cdot v_\theta(x_\nu, \nu, c_t),
    \qquad x_{\nu=0} \sim \mathcal{N}(0,I), \qquad \nu: 0 \to 1.
\end{equation}

\paragraph{Action Mapping.}
We formulate actions as future states, which are more consistent across embodiments than raw control commands, easing cross-platform transfer. However, a single shared output head cannot account for differences in kinematics, control interfaces, and workspace constraints across platforms. We therefore introduce Action Mapping Module, a lightweight adapter that minimizes the embodiment gap and enables deployability across diverse robot platforms.

Architecturally, we initialize it from the last five layers of the action heads. It consumes concatenated embeddings from the low-level VLM and action heads, and outputs robot-specific commands. Each deployed robot uses its own adapter; optional MLPs encode joint/workspace limits to respect physical constraints.

\subsection{Training}

Training proceeds in two phases. First, the Process-Reward Operator (PRO) is pretrained in isolation on real executed trajectories from industrial deployment data, where ground-truth progress, risk, and termination signals are available from operational telemetry. The PRO supervision terms
$\mathcal{L}_{\text{progress}}$, $\mathcal{L}_{\text{risk}}$, and
$\mathcal{L}_{\text{term}}$ (Eqs.~\ref{eq:progress}--\ref{eq:term}) 
are optimized during this stage only and do not enter the main training objective. PRO therefore learns directly from deployment telemetry, independently of policy updates.

Once PRO produces stable signals, its parameters are frozen and it serves as a fixed scoring module in the planning loop. The world model and action heads are then trained jointly on the composite objective in Eq.~\ref{eq:total_loss} with PRO providing advantage supervision $I_t$ to the action heads without receiving gradient updates.

We follow the knowledge insulation scheme~\cite{driess2025knowledgeinsulatingvisionlanguageactionmodels}: in stage one, gradients from the world model and action heads are blocked from the pretrained VLM backbone; in stage two, all components except the frozen PRO are jointly optimized end-to-end.

The planning budget $k$ and horizons $H_{\text{wm}}$, $H_{\text{act}}$ are inference-time hyperparameters and do not affect training.

\subsection{Cross-Embodiment Design}
Across single-arm pick-and-place, dual-arm item sorting, screw sorting, and shoebox unpacking, Cortex 2.0 runs the same planning loop: generate $k$ visual-latent rollouts, score them, and commit to the best trajectory. Because planning operates in visual space, it generalizes across tasks and robot embodiments without modification. Embodiment-specific adaptation is handled entirely by the action heads (Section~\ref{sec:vla}).
\section{Dataset Composition}
Cortex~2.0 is trained on a heterogeneous corpus combining proprietary 
deployment data, targeted teleoperation demonstrations, open-source 
cross-embodiment datasets, and synthetic simulation data. 
Table~\ref{tab:dataset} summarizes the composition.

\begin{table}[h]
\centering
\caption{Dataset composition. F/T = force--torque sensor; Proprio = 
proprioceptive state (joint positions, velocities, torques).}
\label{tab:dataset}
\resizebox{\columnwidth}{!}{%
\begin{tabular}{lrrlp{4.5cm}}
\toprule
Source & Episodes & Hours & Task families & Sensors \\
\midrule
Deployment   & $>$10M & $>$25k & Pick, pack, unpack, sort, returns, kitting, assembly, navigation & RGB, F/T, vacuum, proprio \\
Teleoperation & ${\sim}$40k & ${\sim}$400 & Pick, pack, unpack, sort, returns, kitting, navigation & RGB, F/T, proprio \\
Open-source  & ${\sim}$970k & ${\sim}$2k & Cross-embodiment diverse & RGB, proprio \\
Synthetic    & ${\sim}$20k & --- & Pick, sort, table-top rearrangement & RGB, proprio \\
\bottomrule
\end{tabular}%
}
\end{table}

\subsection{Real-World Deployment Data}
\begin{figure*}[htbp]
\centering
\setlength{\tabcolsep}{1pt}
\begin{tabular}{ccc}
\includegraphics[width=0.19\linewidth]{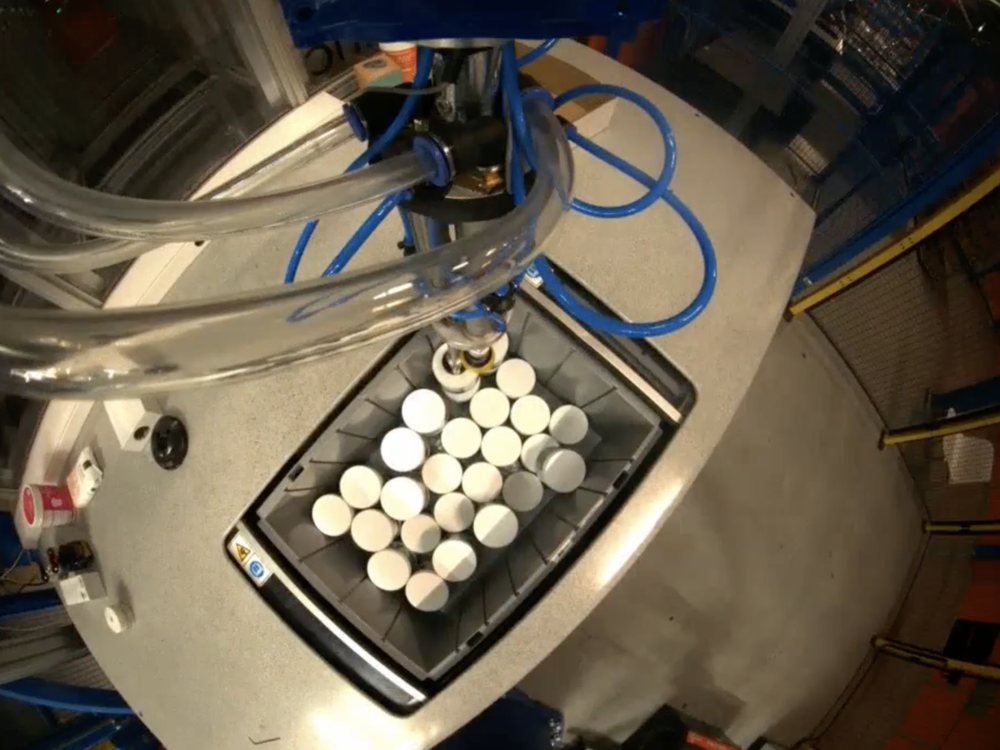} &
\includegraphics[width=0.19\linewidth]{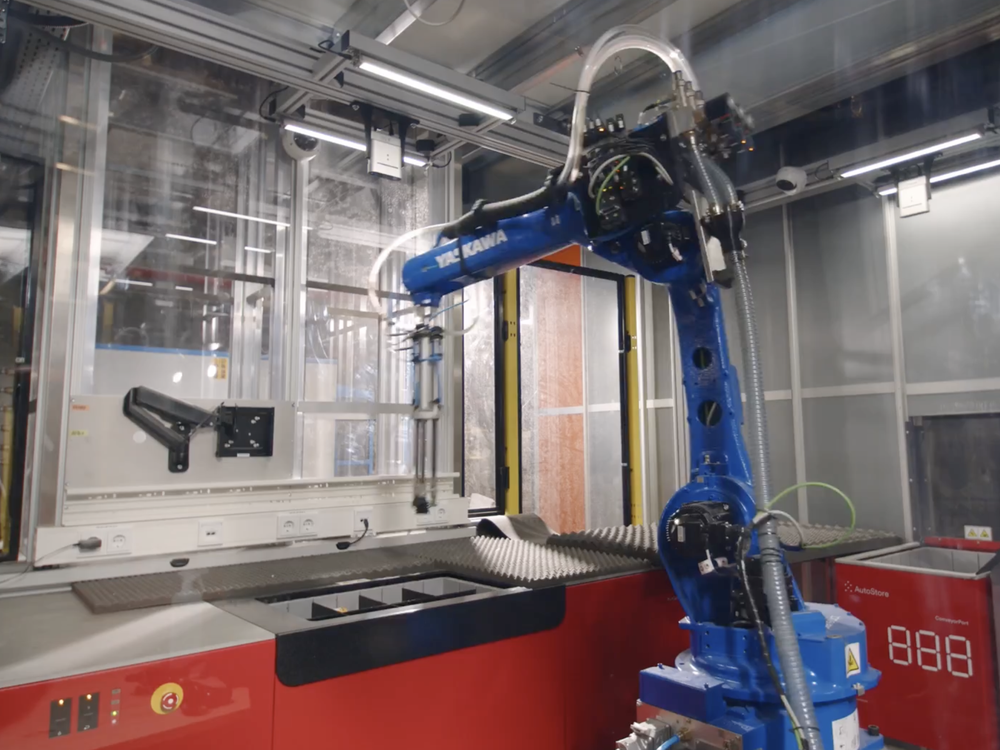} &
\includegraphics[width=0.19\linewidth]{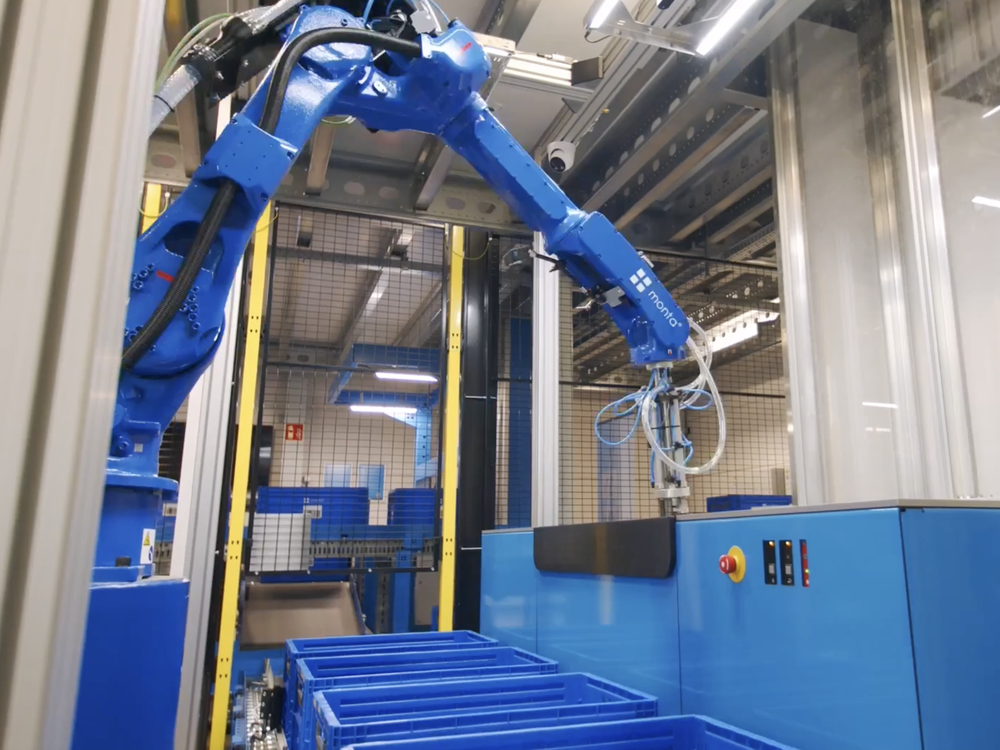} \\
\includegraphics[width=0.19\linewidth]{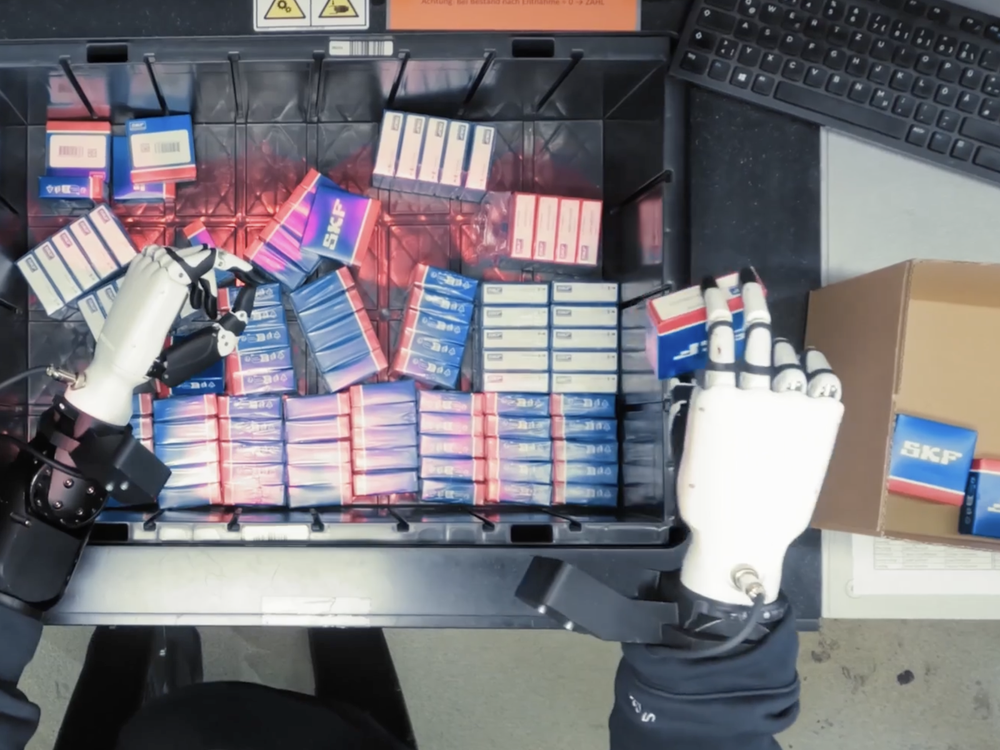} &
\includegraphics[width=0.19\linewidth]{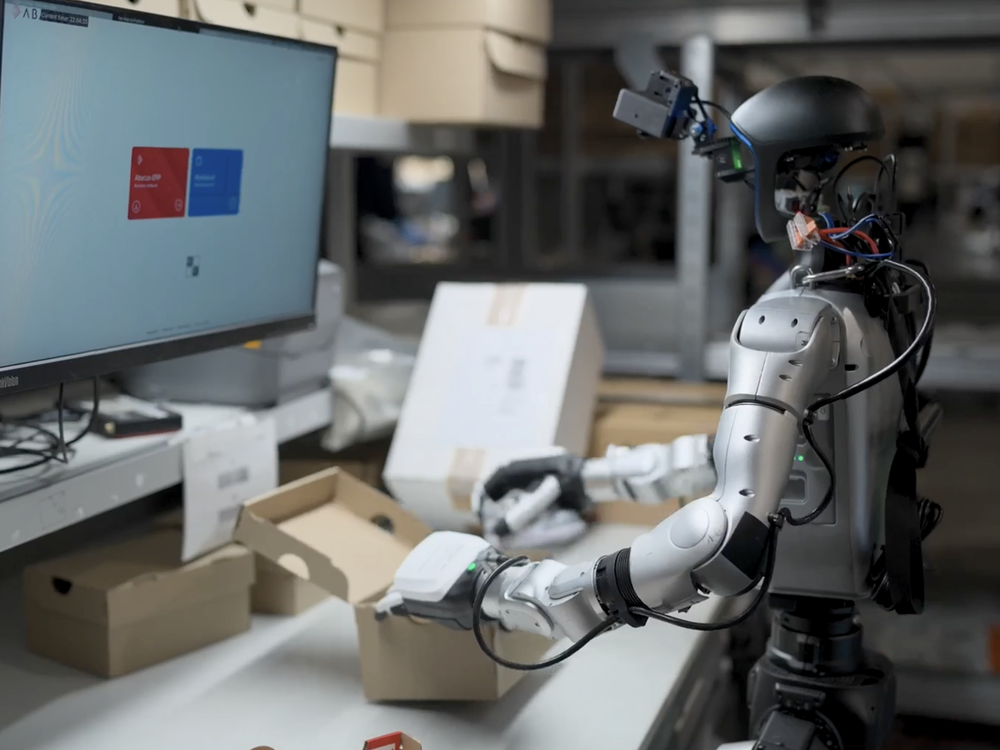} &
\includegraphics[width=0.19\linewidth]{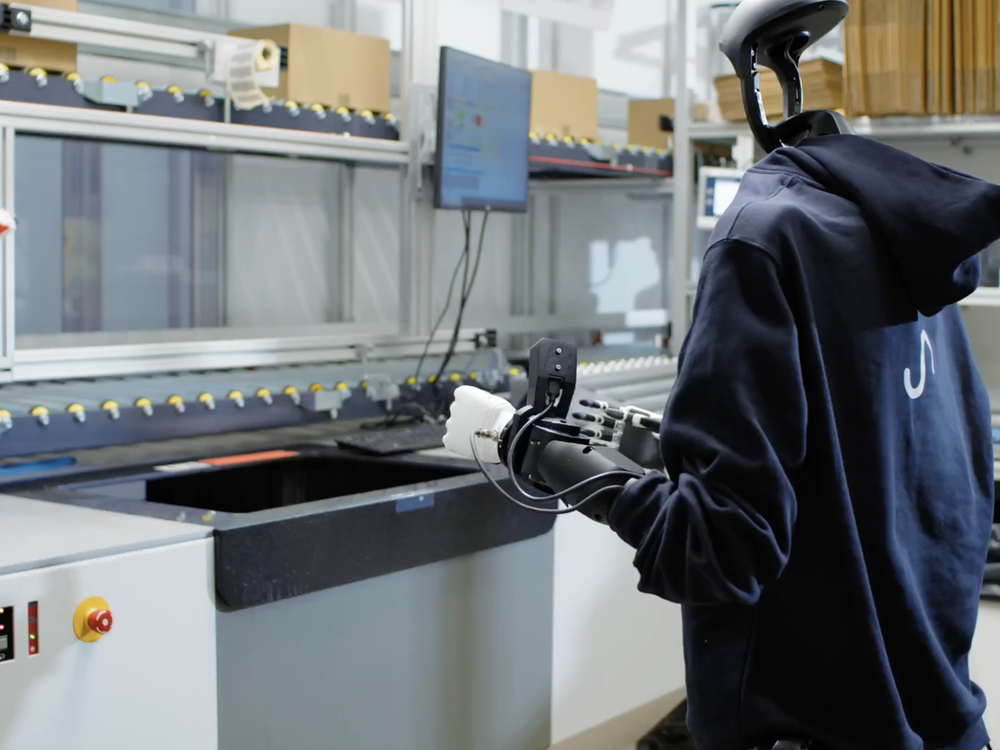} \\
\includegraphics[width=0.19\linewidth]{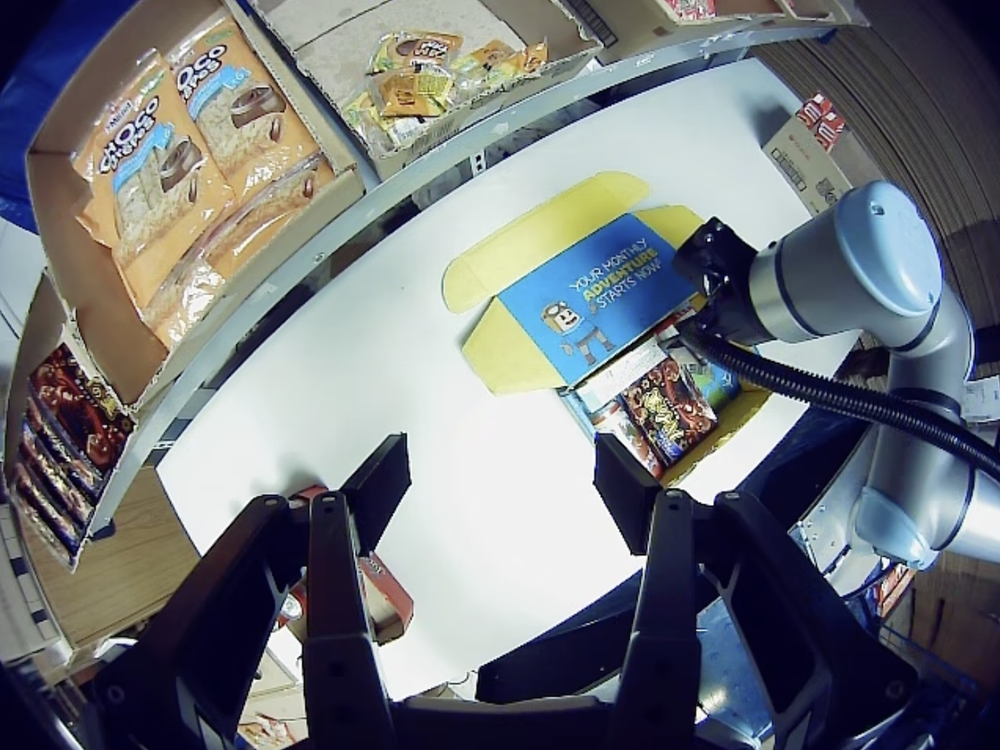} &
\includegraphics[width=0.19\linewidth]{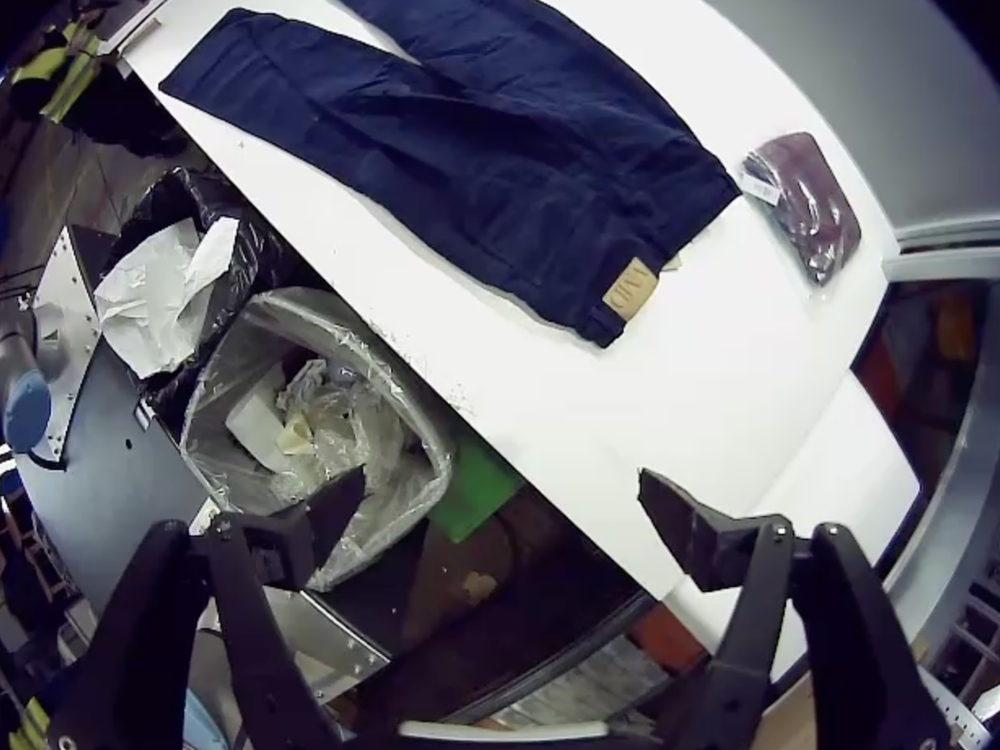} &
\includegraphics[width=0.19\linewidth]{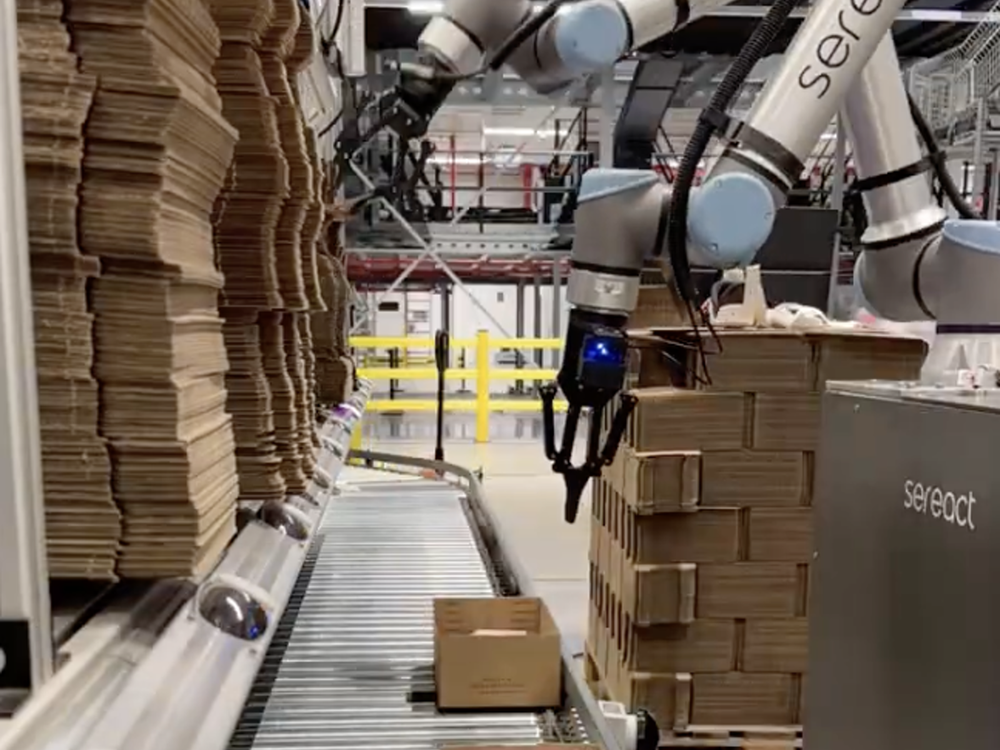} \\
\end{tabular}
\caption{Real Production Robotic Data.}
\label{fig:real-production}
\end{figure*}

Our fleet has accumulated over 500 million manipulation interactions across warehouse deployments, collected continuously at 30\,Hz from operational robotic systems (Figure~\ref{fig:real-production}). All sensory observations including RGB images from wrist cameras, proprioceptive signals, and operational telemetry including force–torque, vacuum pressure, and contact signals, are recorded simultaneously. Telemetry directly supervises PRO; visual observations train the world model. Cortex 2.0 is trained on a curated subset of 10 million interactions sampled to preserve task diversity and coverage of failure modes. As fleet size grows and training subset scales, Cortex 2.0 benefits from increased diversity of states and execution contexts, leading to compounding improvements in reward quality and downstream policy performance.

\subsection{Open-Source and Synthetic Data}
\begin{figure*}[htbp]
\centering
\setlength{\tabcolsep}{1pt}
\begin{tabular}{cccc}
\includegraphics[width=0.19\linewidth]{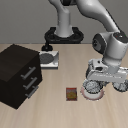} &
\includegraphics[width=0.19\linewidth]{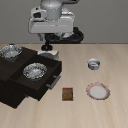} &
\includegraphics[width=0.19\linewidth]{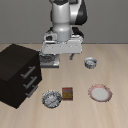} &
\includegraphics[width=0.19\linewidth]{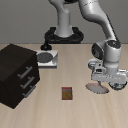} \\
\includegraphics[width=0.19\linewidth]{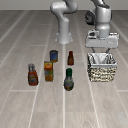} &
\includegraphics[width=0.19\linewidth]{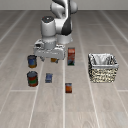} &
\includegraphics[width=0.19\linewidth]{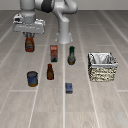} &
\includegraphics[width=0.19\linewidth]{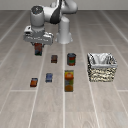} \\
\includegraphics[width=0.19\linewidth]{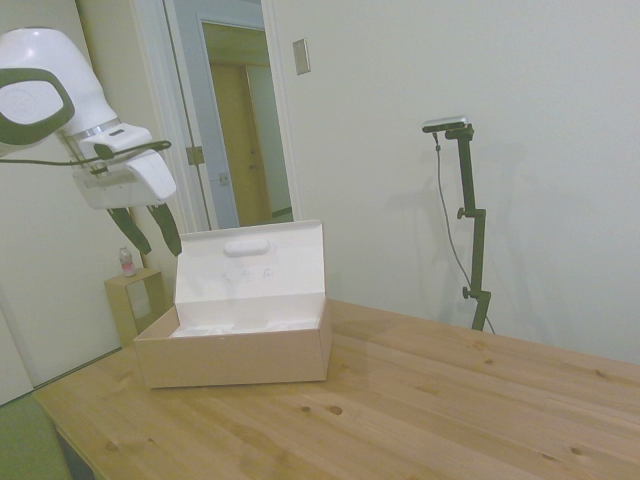} &
\includegraphics[width=0.19\linewidth]{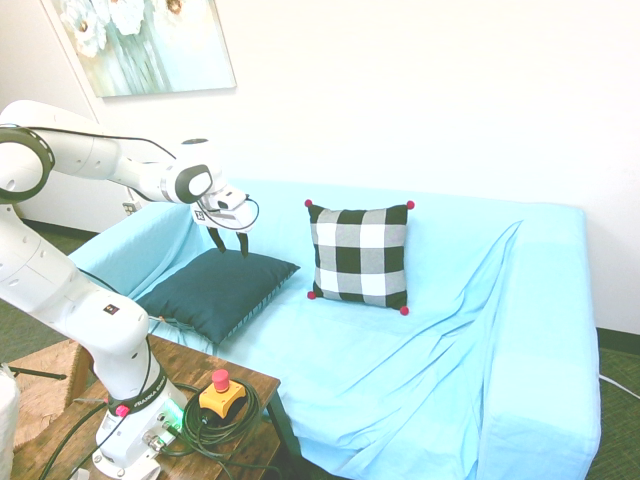} &
\includegraphics[width=0.19\linewidth]{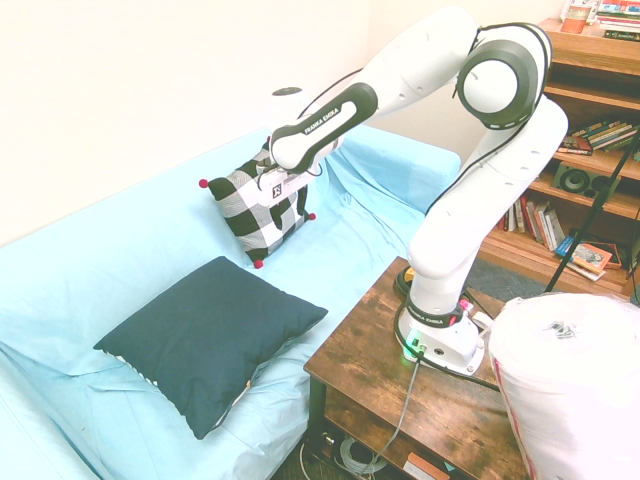} &
\includegraphics[width=0.19\linewidth]{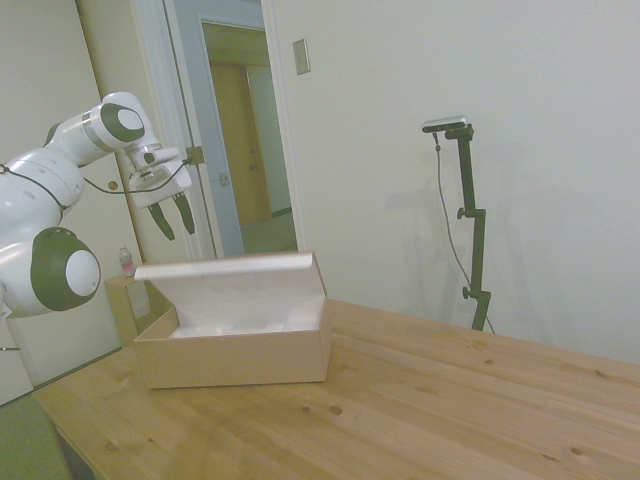} \\
\end{tabular}
\caption{Open Source Datasets.}
\label{fig:open_source}
\end{figure*}

\begin{figure*}[htbp]
\centering
\begin{subfigure}[t]{\textwidth}
\centering
\includegraphics[width=0.19\linewidth]{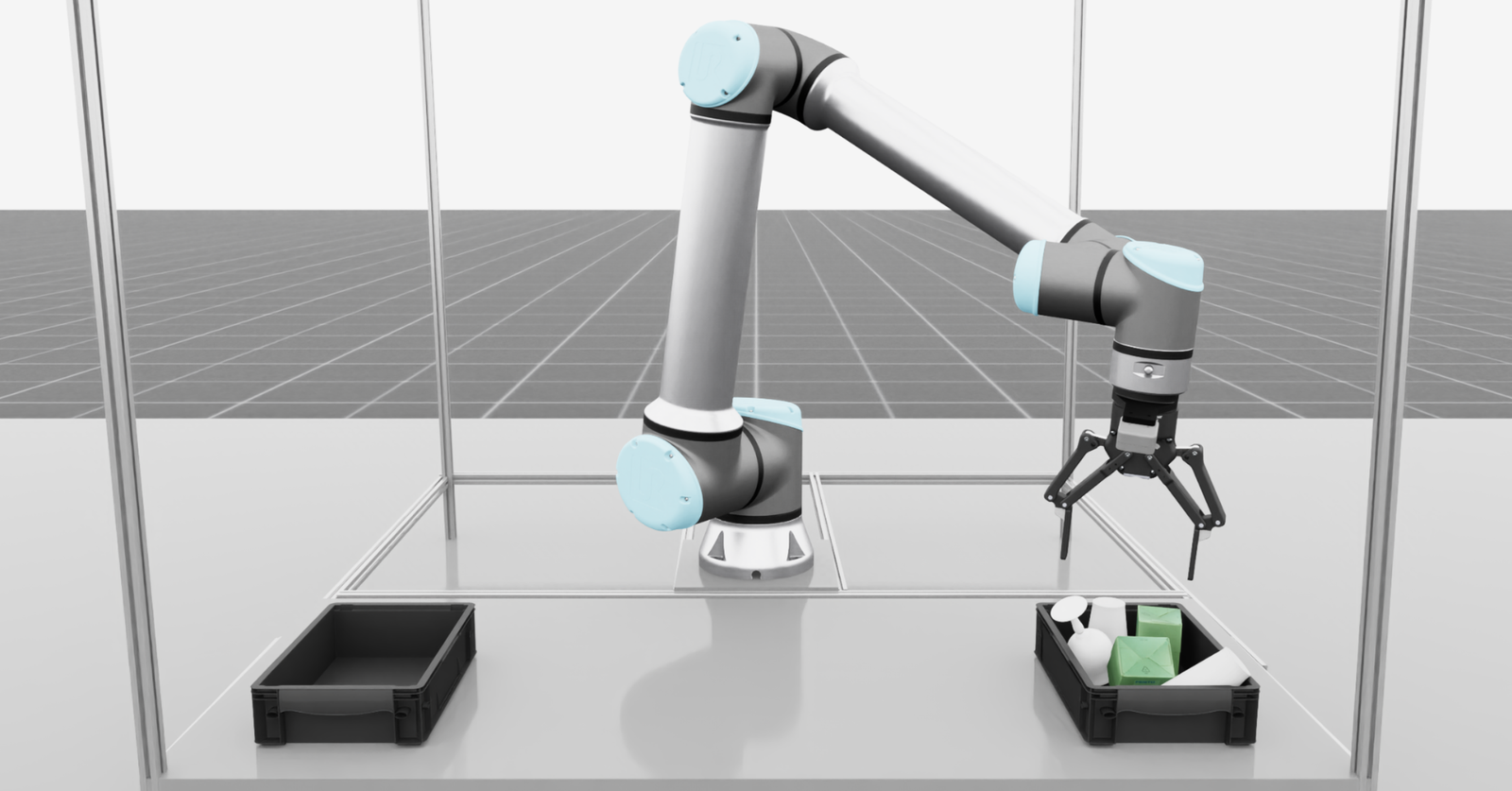}
\includegraphics[width=0.19\linewidth]{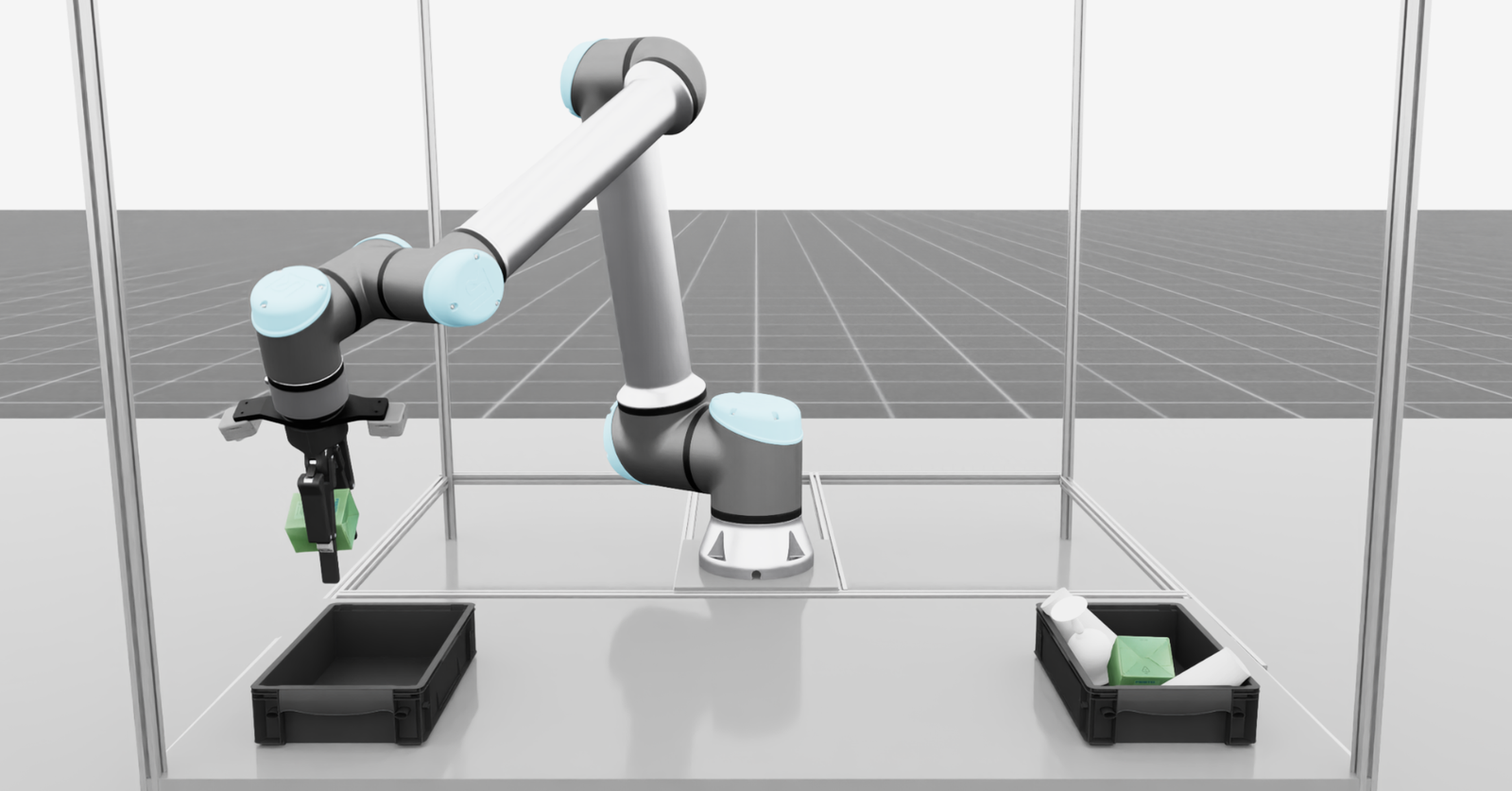}
\includegraphics[width=0.19\linewidth]{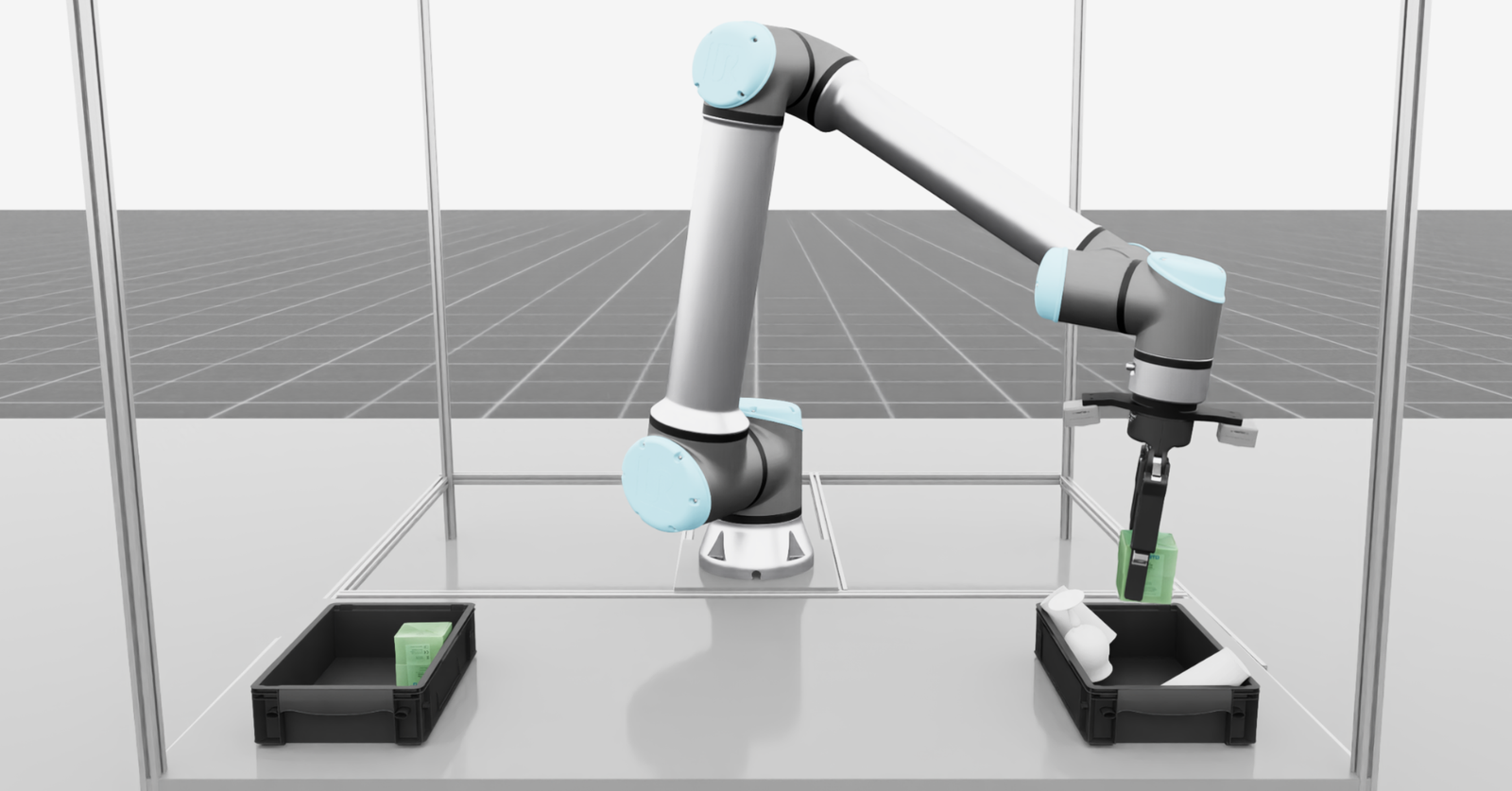}
\includegraphics[width=0.19\linewidth]{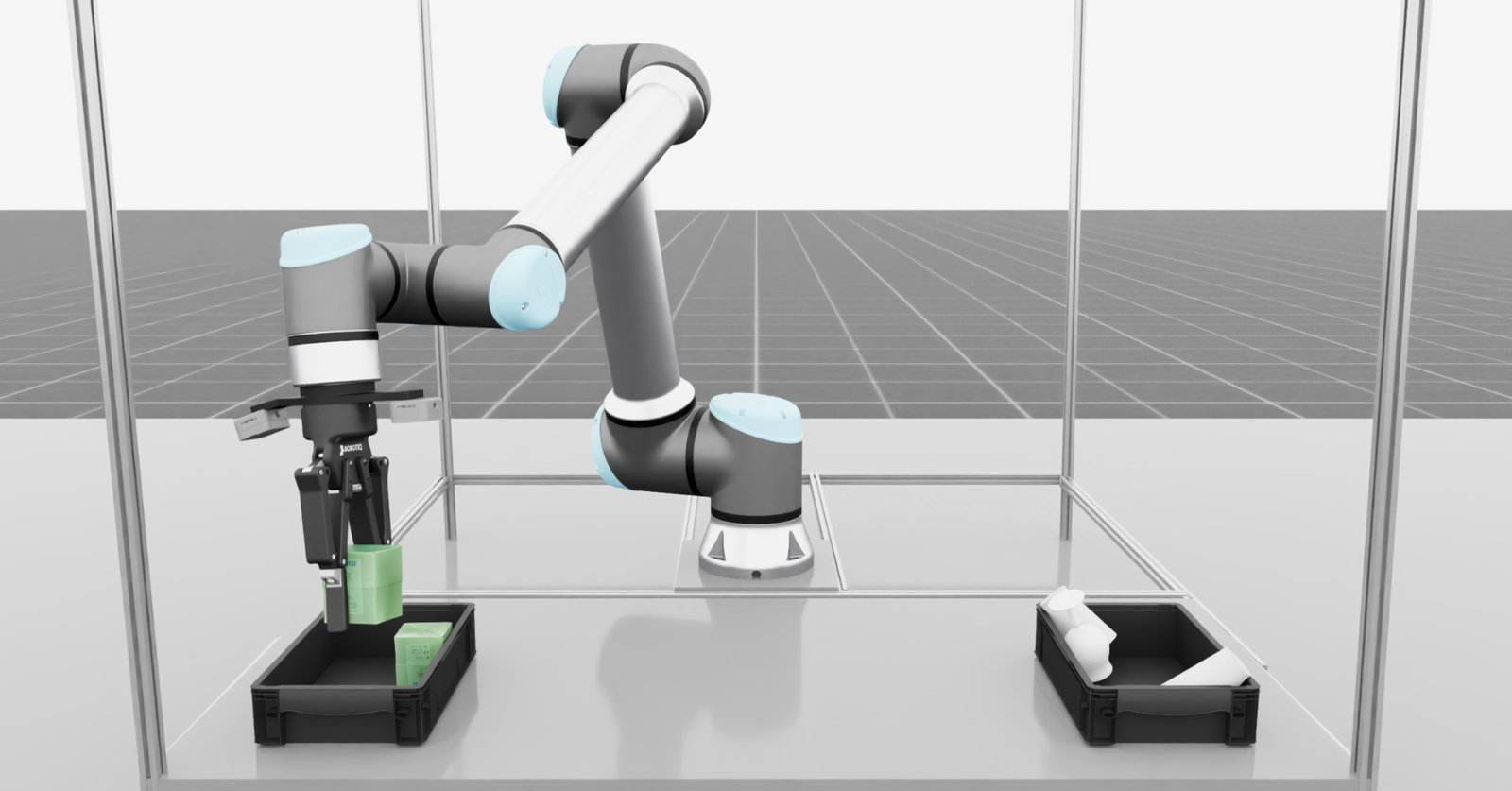}
\caption{Pick and Place - Single Arm}
\label{fig:sim-pick-place}
\end{subfigure}

\begin{subfigure}[t]{\textwidth}
\centering
\includegraphics[width=0.19\linewidth]{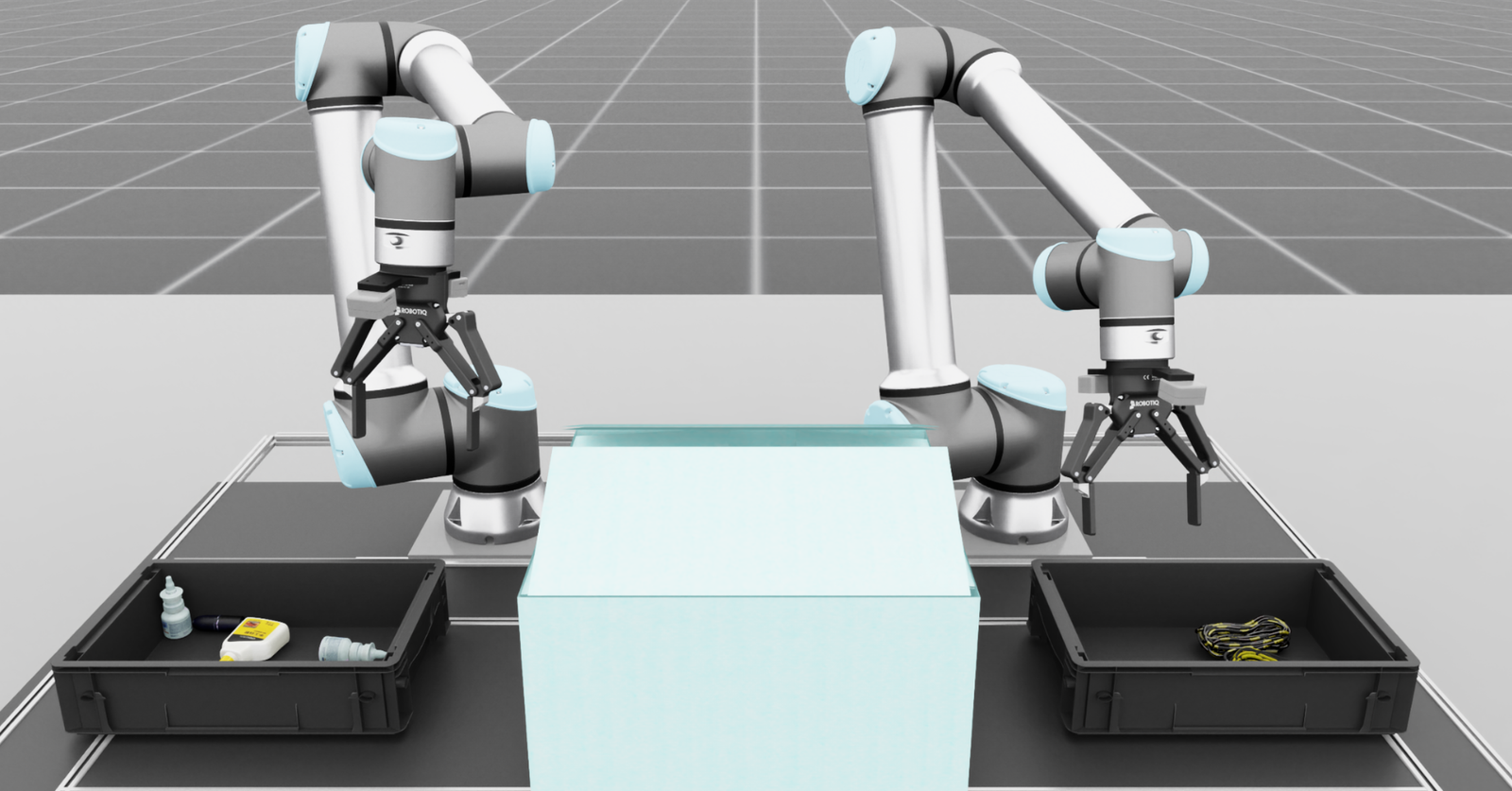}
\includegraphics[width=0.19\linewidth]{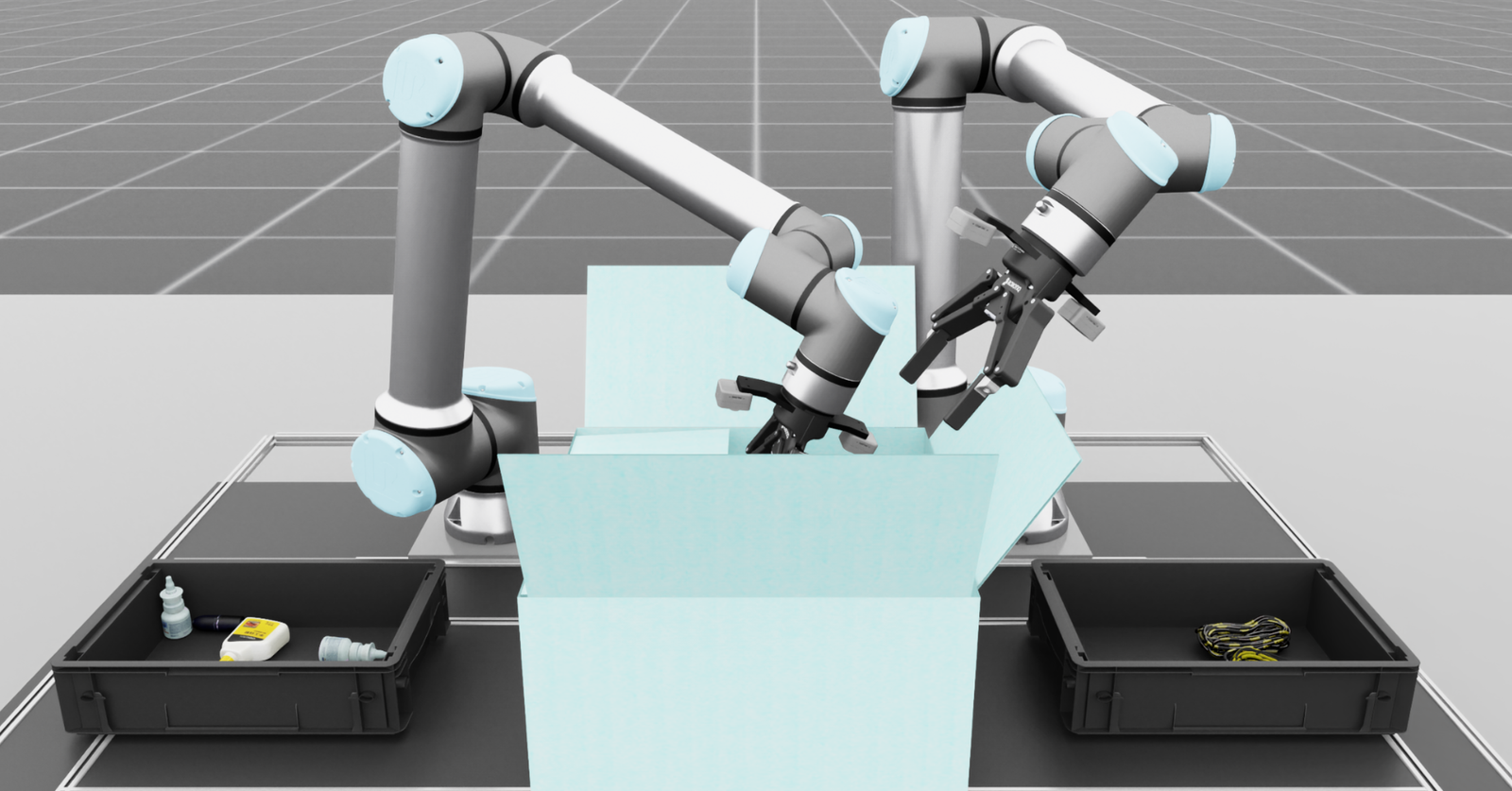}
\includegraphics[width=0.19\linewidth]{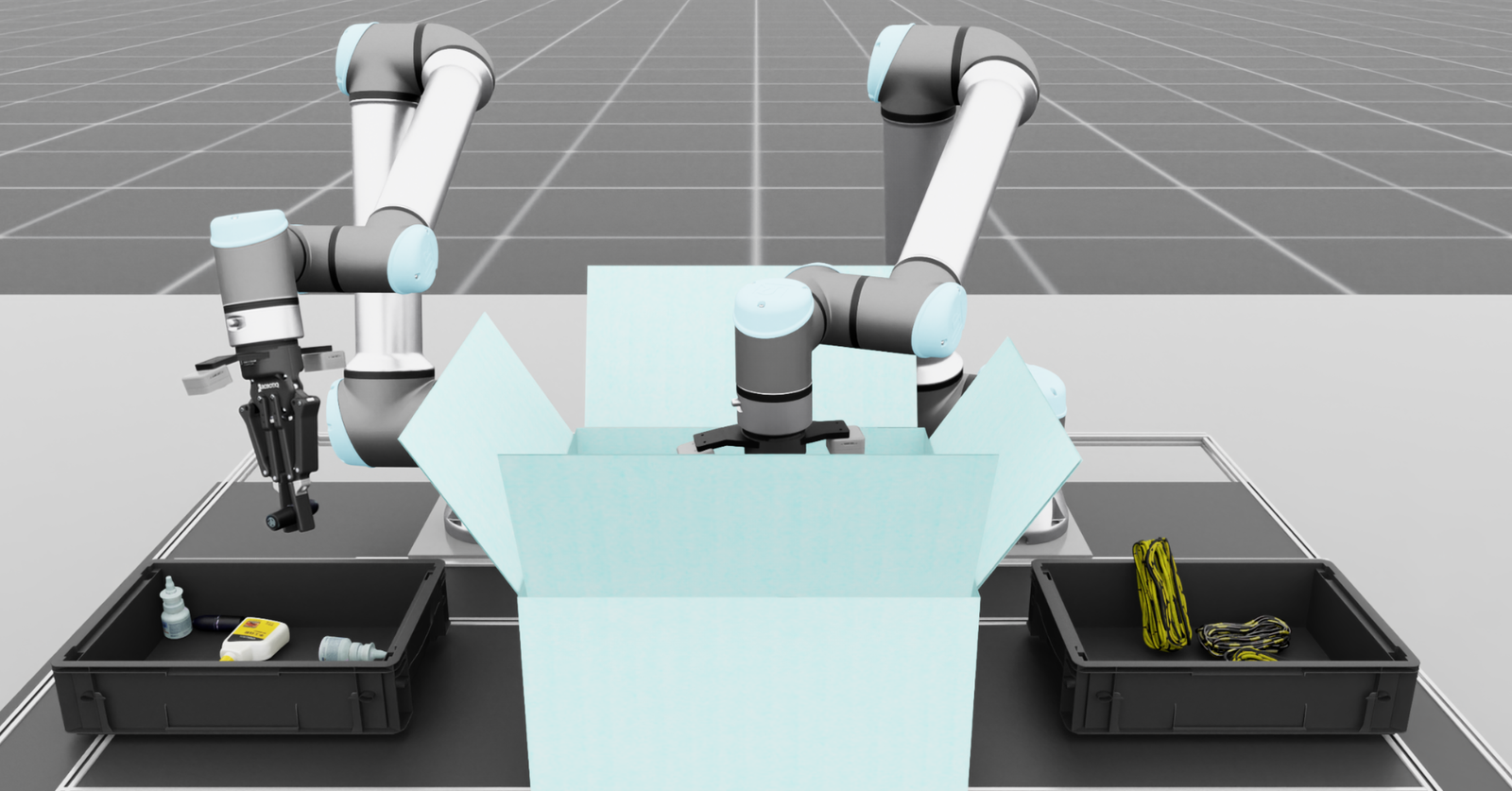}
\includegraphics[width=0.19\linewidth]{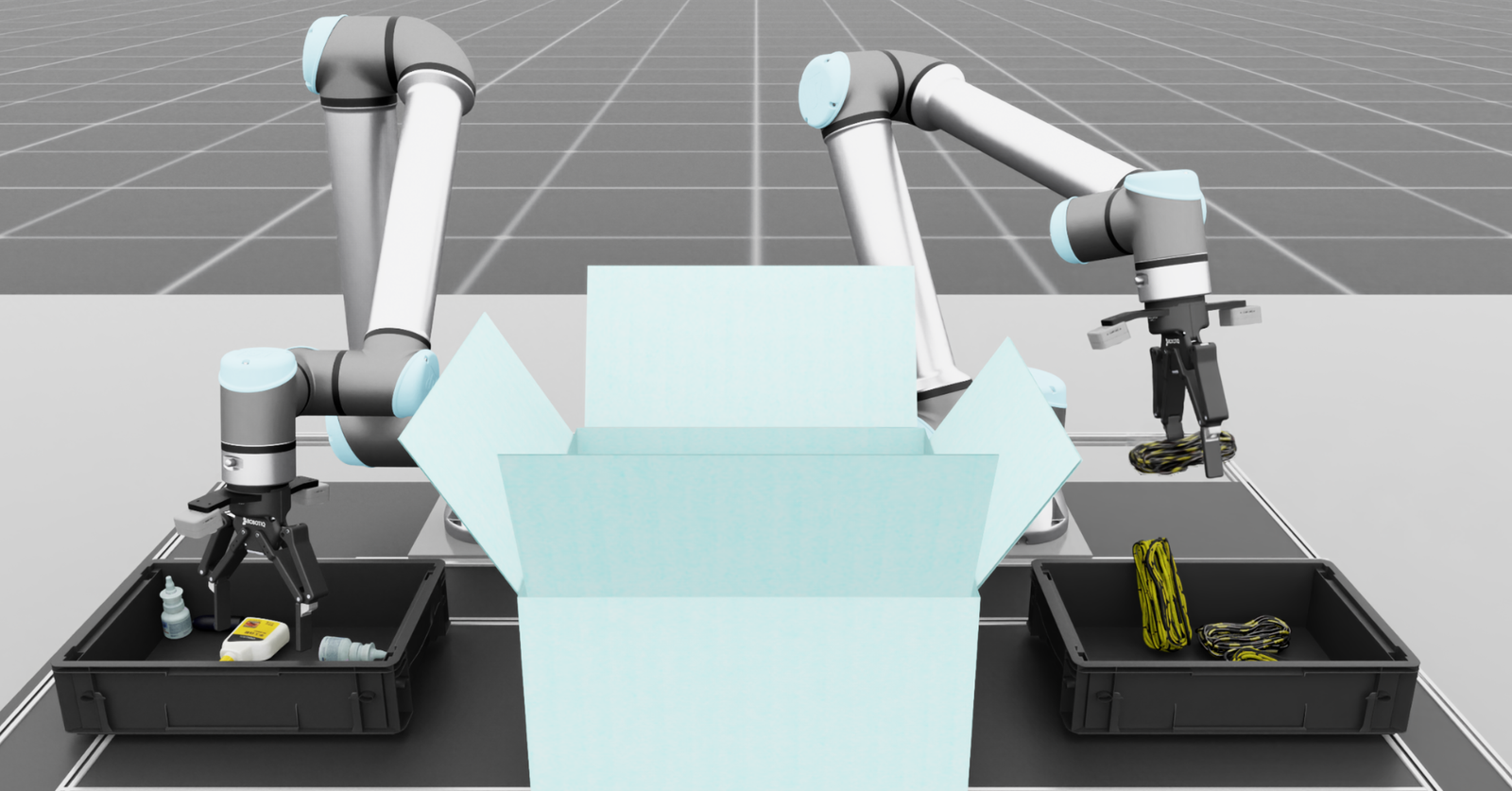}
\caption{Sorting and Packaging}
\label{fig:dual-sorting-and-packaging}
\end{subfigure}

\begin{subfigure}[t]{\textwidth}
\centering
\includegraphics[width=0.19\linewidth]{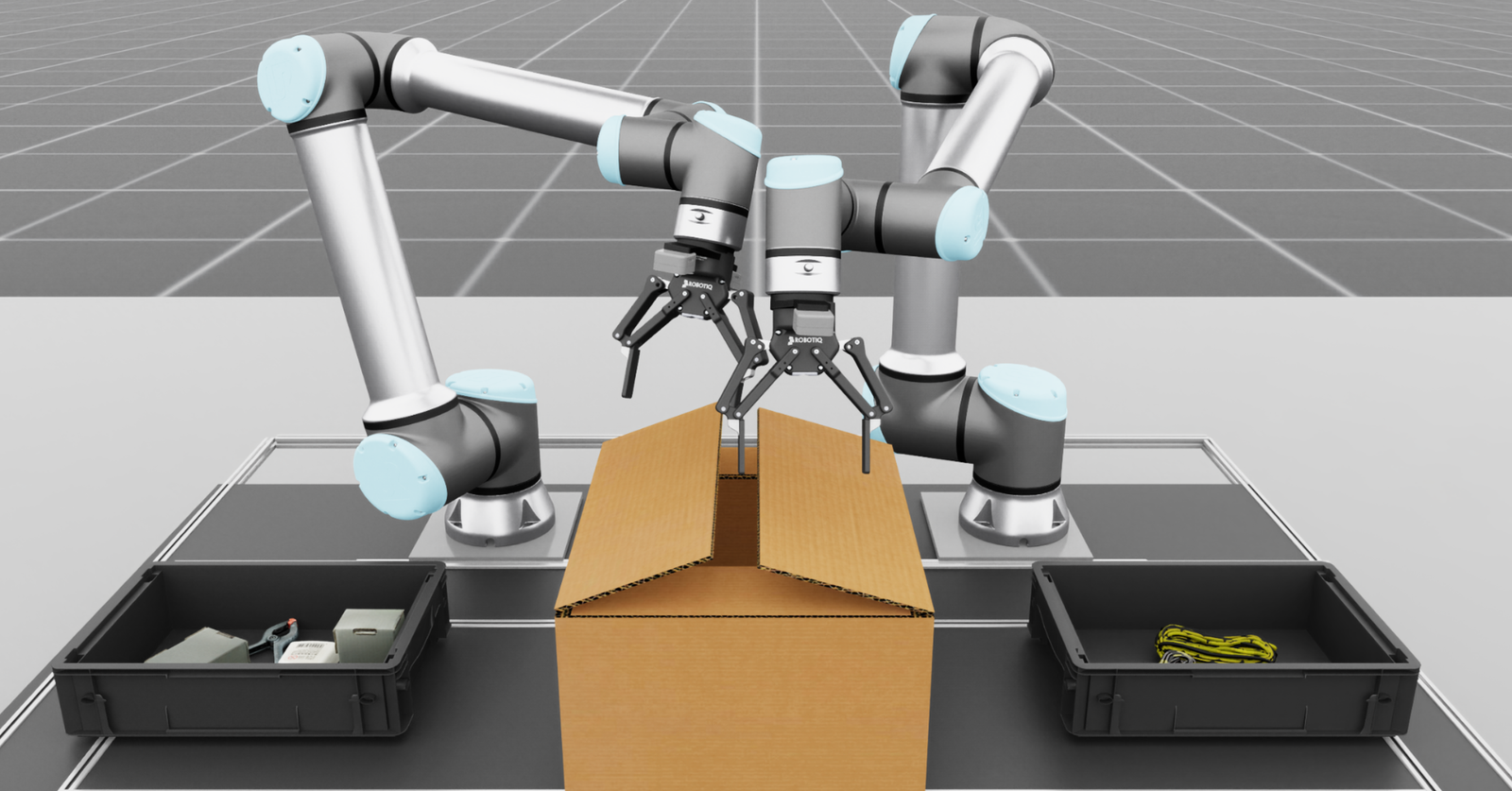}
\includegraphics[width=0.19\linewidth]{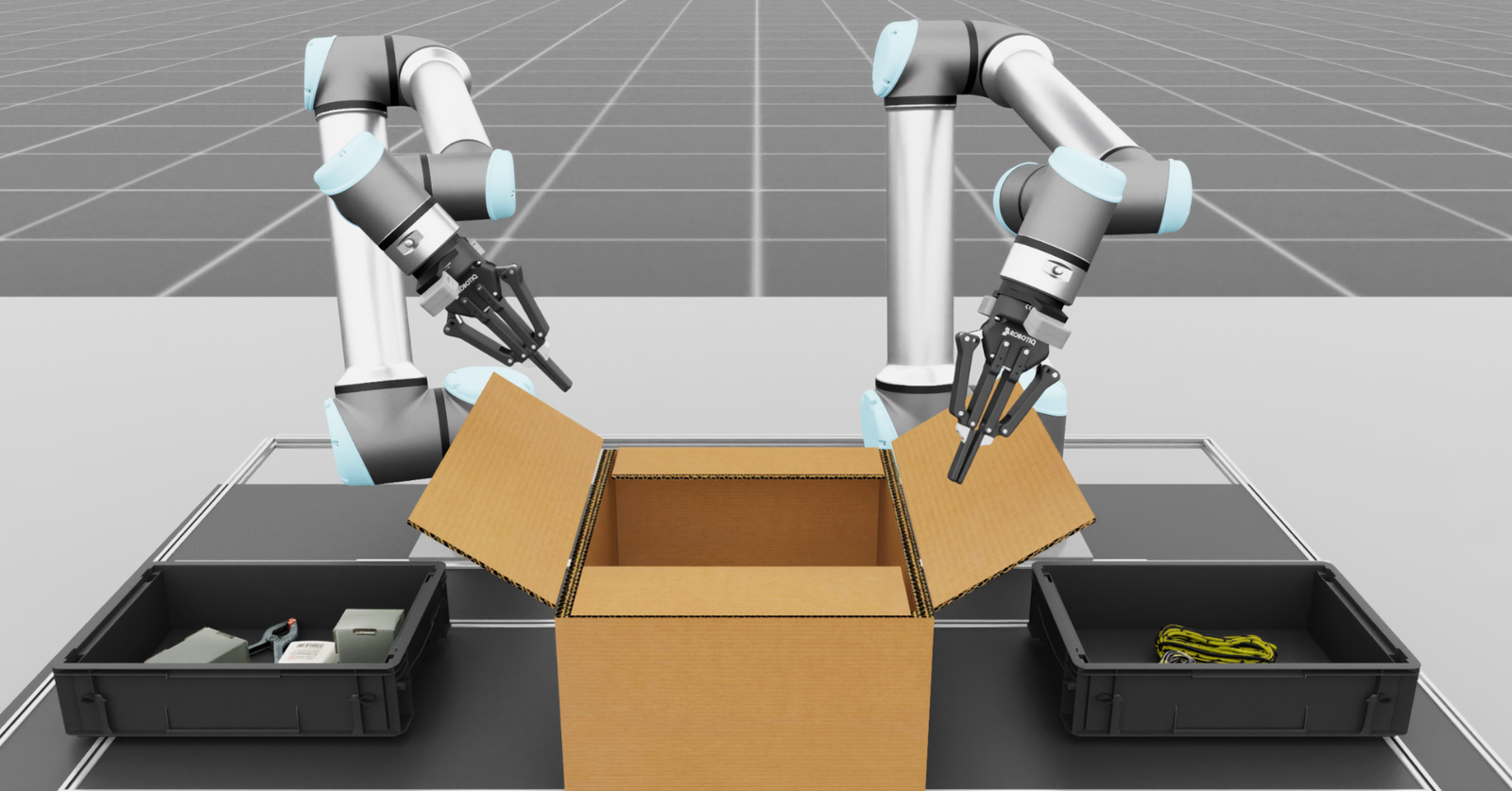}
\includegraphics[width=0.19\linewidth]{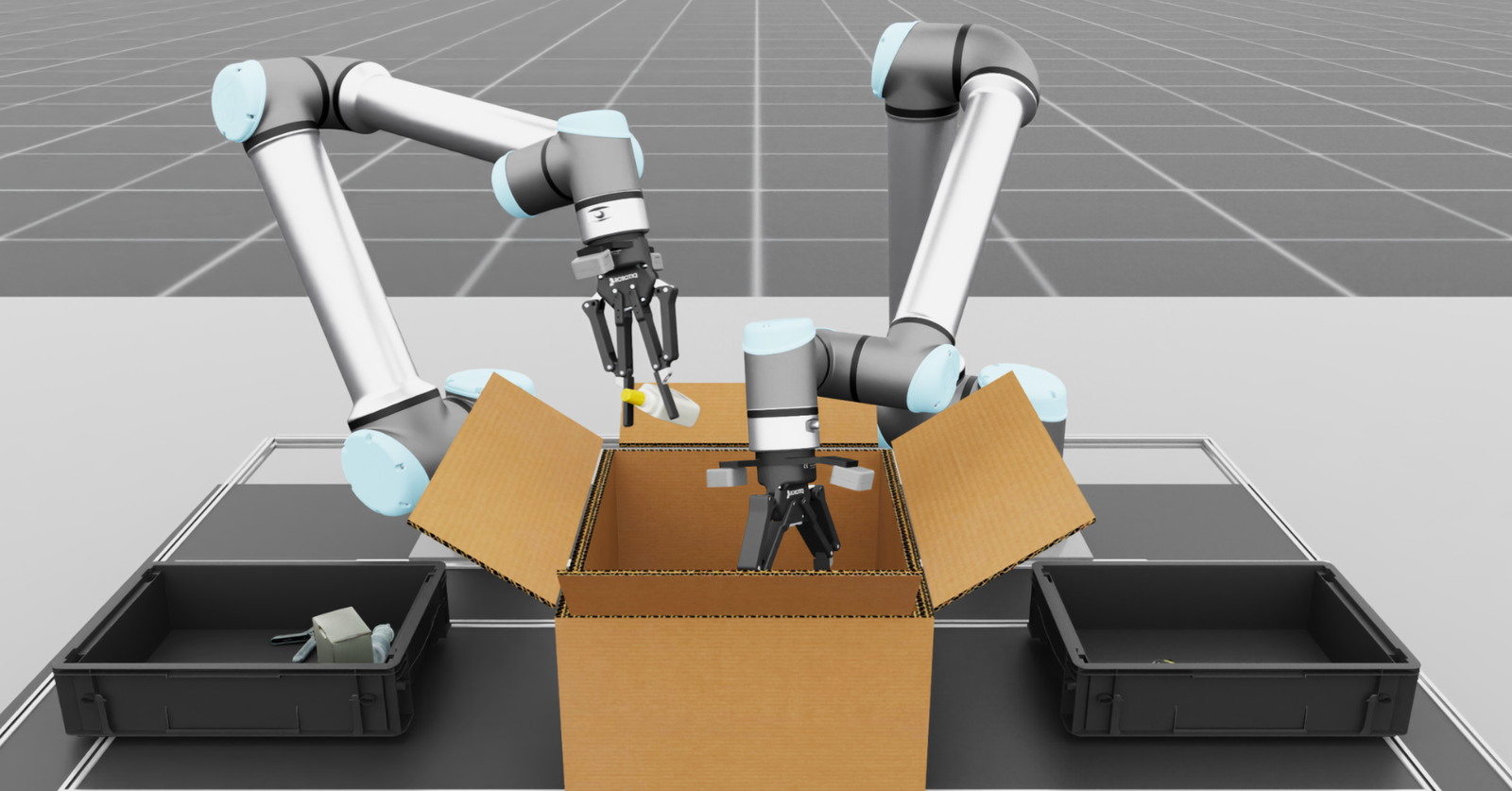}
\includegraphics[width=0.19\linewidth]{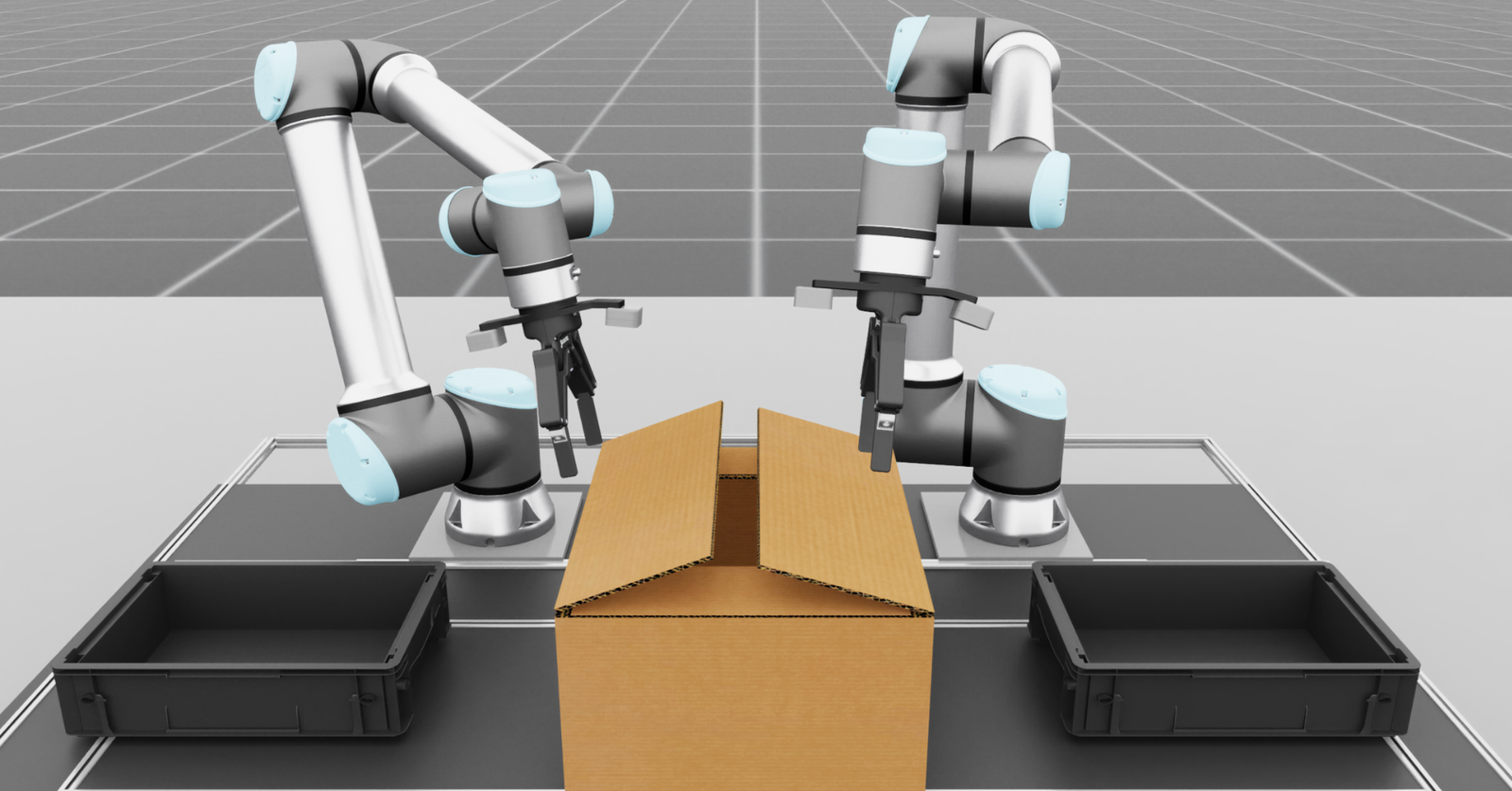}
\caption{Pick and Packaging}
\label{fig:dual-pick-and-packaging}
\end{subfigure}

\begin{subfigure}[t]{\textwidth}
\centering
\includegraphics[width=0.19\linewidth]{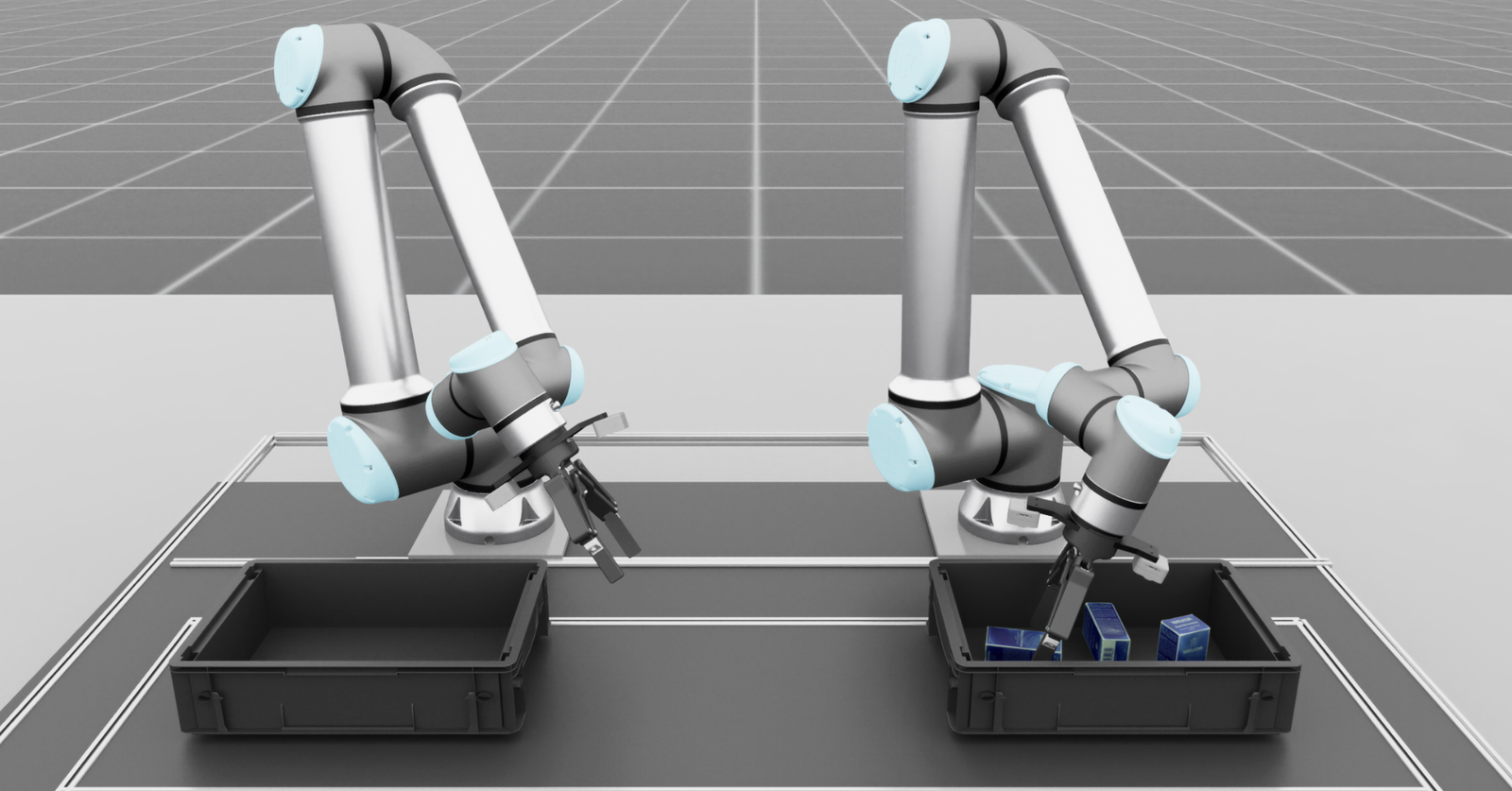}
\includegraphics[width=0.19\linewidth]{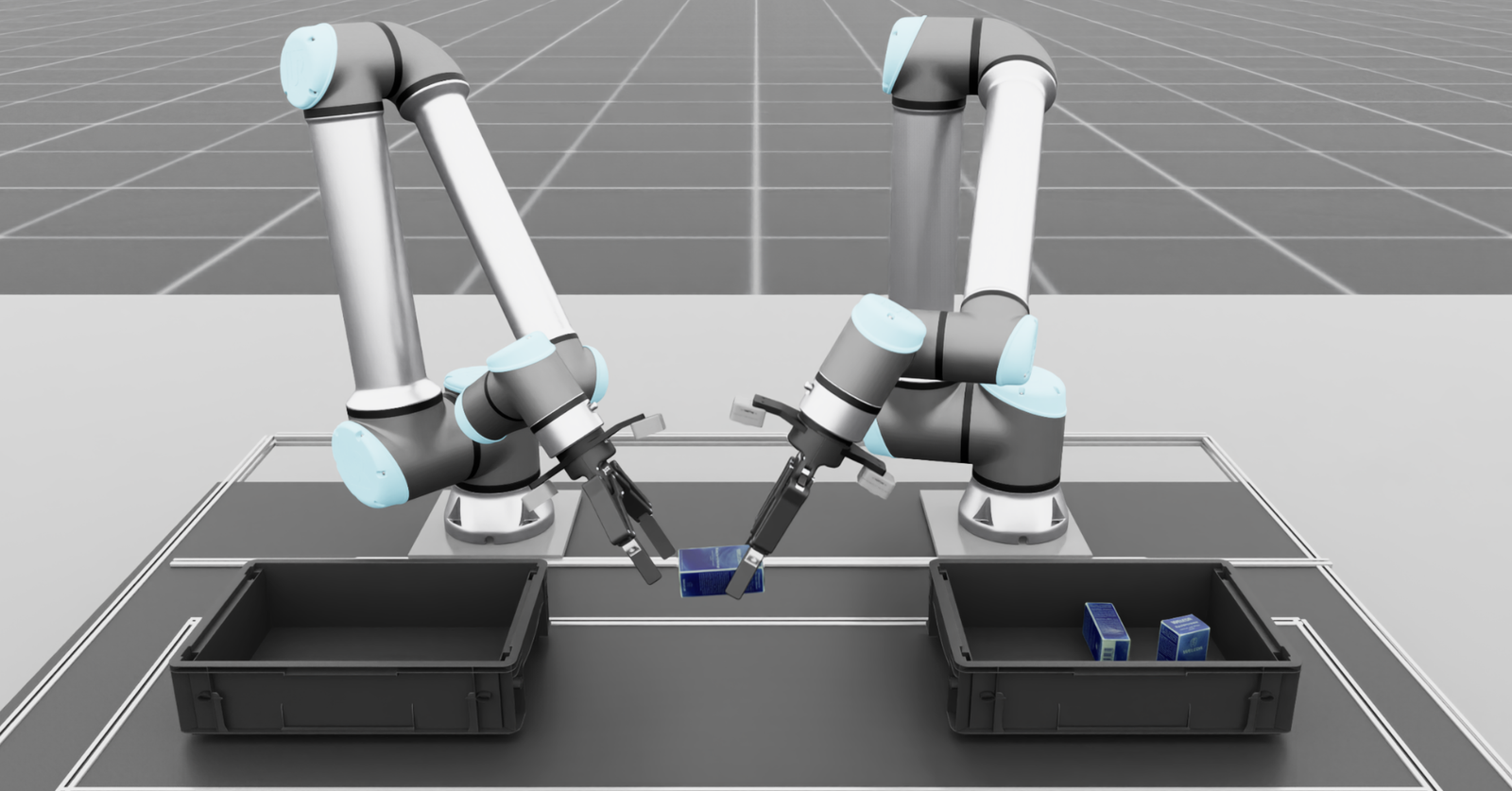}
\includegraphics[width=0.19\linewidth]{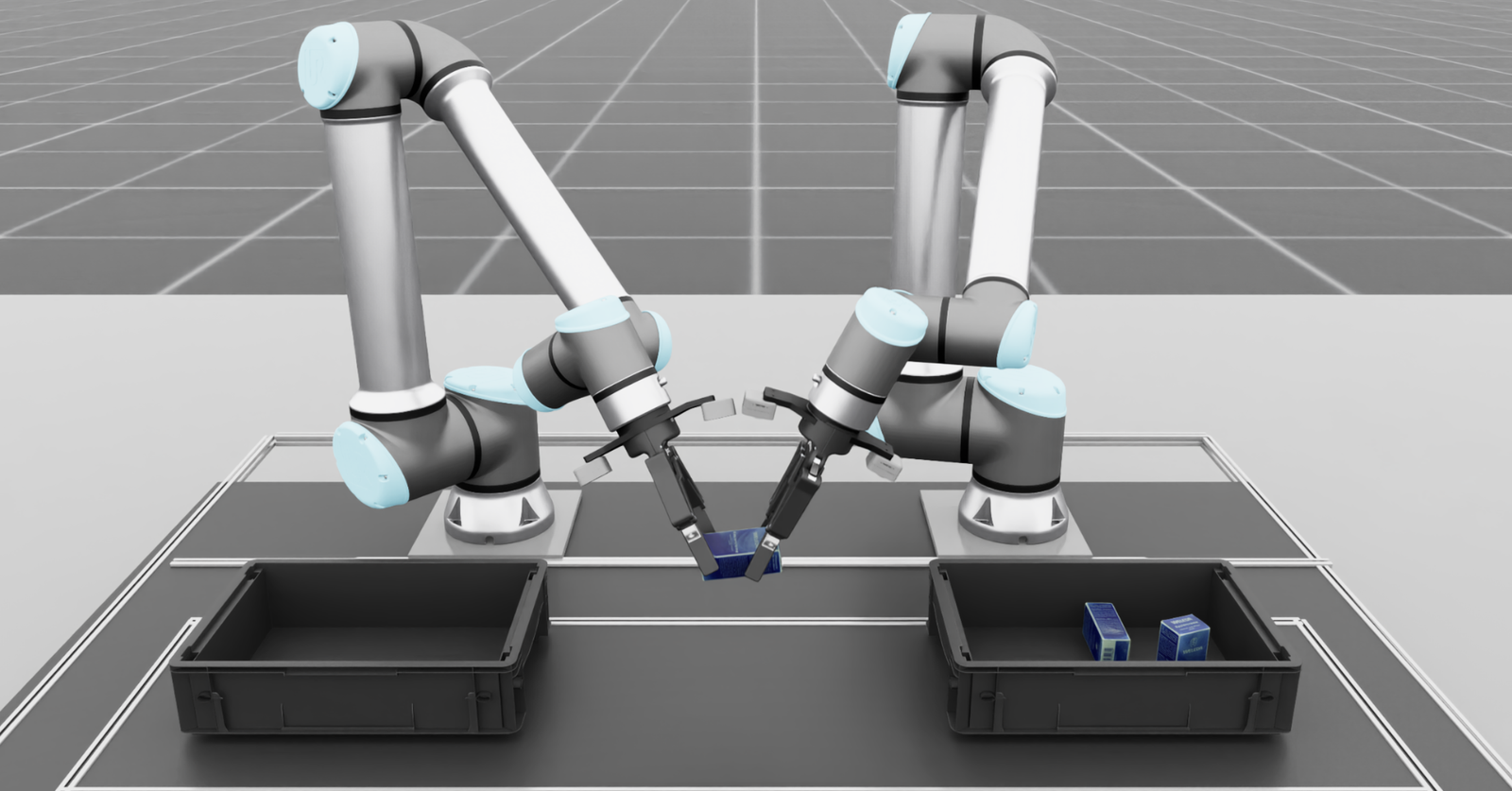}
\includegraphics[width=0.19\linewidth]{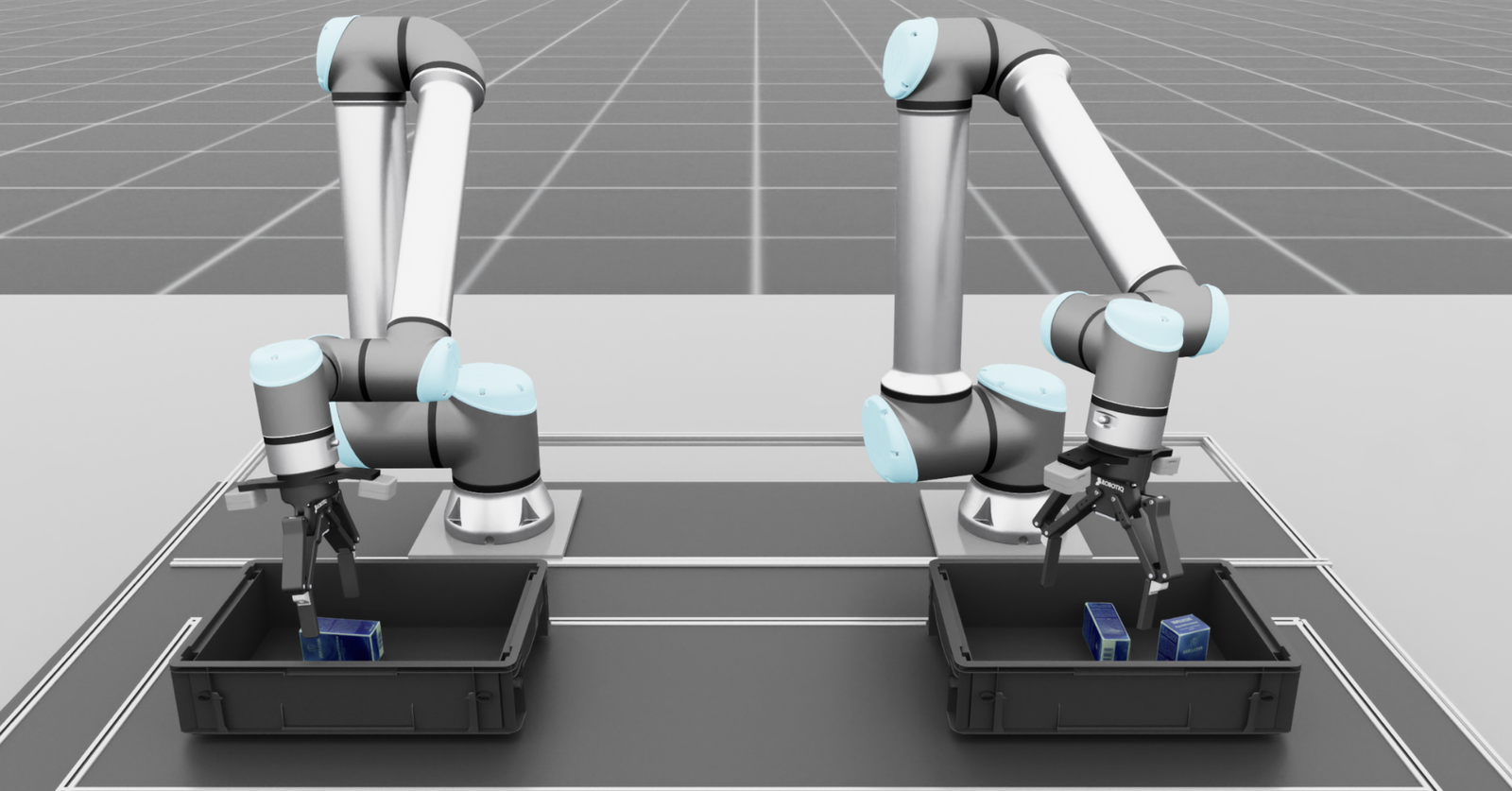}
\caption{Transport}
\label{fig:dual-transport}
\end{subfigure}
\caption{Synthetic Isaac Sim data -- Dual Robotic Arms.}
\label{fig:sim-main}
\end{figure*}

To broaden embodiment and task diversity, we incorporate large-scale publicly available robot datasets (Figure~\ref{fig:open_source}) including components of Open X-Embodiment \cite{OXE2024}, BridgeData V2 \cite{Bridgev22023}, and DROID \cite{khazatsky2024droid}. Synthetic data generated in simulation using the RoboCasa framework \cite{robocasa2024} augments environmental diversity and trajectory variability across a broad set of manipulation tasks~\ref{fig:sim-main}. Together, these components form a unified corpus that bridges real and simulated data, supporting both fine-grained control learning and high-level reasoning across embodiments.

\subsection{In-House Teleoperation Data}

A core component of our training corpus is in-house data collection across a wide range of warehouse operation tasks under both single- and dual-arm configurations. We collect through complementary channels.
Egocentric recordings, captured from head-mounted cameras worn during normal workflow, yield data at high scale with minimal overhead.
Full-body human suite recordings capture whole-body motion via inertial and optical motion capture, essential for tasks requiring coordinated reach, bimanual handling and posture-dependent manipulation. 
Handheld gripper demonstrations use an instrumented end-effector, producing trajectories whose sensory and kinematic distribution closely matches the deployed robot. This channel is particularly valuable for contact-rich tasks such as screw sorting, kitting, and returns handling.
Teleoperation allows operators to remotely control the target robot in real time, preserving exact embodiment dynamics and enabling fast episode collection for long-horizon and deformable-object tasks.
We record synchronized visual observations, gripper force feedback, and full robot state. Our teleoperation framework maps hand motions to robot TCP trajectories via inverse kinematics, achieving latencies at the 10,ms level and sub-centimeter replay precision.

\subsection{World Model Pretraining Data}
The world model is pretrained on all visual data prior to fine-tuning on deployment recordings. This approach enables general physical priors to be acquired from diverse video sources before being adapted to deployment-specific manipulation dynamics. Deployment data collected from live operations feeds back into training continuously, creating a closed-loop improvement cycle where better world models enable more accurate planning, which improves execution quality, which in turn generates cleaner training signal.

\section{Experiments}

\subsection{Experimental Setup}

We evaluate Cortex~2.0 against state-of-the-art open-source visuomotor policies on a single-arm and a dual-arm manipulation platform, each equipped with a Universal Robot arm and a parallel gripper. Visual observations come from wrist-mounted cameras on each end-effector. Actions are executed at 30\,Hz.

\subsubsection{Baselines}

We compare against three policies: $\pi_{0.5}$~\cite{pi05} operating in absolute joint space; Diffusion Policy~\cite{diffusionpolicy} using relative end-effector actions; and RDT-2~\cite{liu2026rdt2}, a diffusion transformer for bimanual manipulation. All models are trained with equivalent computational budgets of 200~GPU hours to ensure fair comparison. The $\pi_{0.5}$ \cite{pi05} policy is trained from the pretrained checkpoint from LeRobot \cite{cadene2024lerobot}. Diffusion Policy, due to its smaller capacity, is trained from scratch for 100,000 steps, with the ResNet-18 backbone initialized from pretrained ImageNet weights. For RDT-2 we use the official implementation~\cite{liu2026rdt2}. Cortex~2.0, Diffusion Policy, and RDT-2 operate in Cartesian space with relative end-effector actions, predicting translational displacements and 6D rotation representations. $\pi_{0.5}$ uses absolute joint-space observations and actions, directly predicting target joint configurations. Table~\ref{tab:models} summarizes the model configurations. For both our model and baselines, we use a action chunk of size 12. The execution of the actions is at 30\,Hz for all the embodiment. 

\begin{table}[h]
\centering
\caption{Baseline model configurations. All models trained with equivalent computational budgets.}
\label{tab:models}
\begin{tabular}{lccc}
\toprule
\textbf{Model} & \textbf{Compute} & \textbf{Obs.\ Space} & \textbf{Action Space} \\
\midrule
Cortex 2.0 (ours)        & 200 GPU-hr & Cartesian       & Rel.\ end-effector \\
Diffusion Policy~\cite{diffusionpolicy} & 200 GPU-hr & Cartesian       & Rel.\ end-effector \\
$\pi_{0.5}$~\cite{pi05}  & 200 GPU-hr & Absolute joint  & Absolute joint \\
RDT-2~\cite{liu2026rdt2}  & 200 GPU-hr & Cartesian       & Rel.\ end-effector \\
\bottomrule
\end{tabular}
\end{table}

\subsubsection{Evaluation Protocol}

Across all experiments we track unrecoverable states where human intervention is required. These are recorded under three conditions:
\begin{enumerate}
    \item \textbf{Safety-critical collisions}: the robot collides with the environment, itself, or the other arm; execution is halted and the system is re-homed before continuing.
    \item \textbf{Persistent control deadlocks}: the policy enters repeated or oscillatory motion without measurable task progress that does not resolve through continued execution.
    \item \textbf{Unrecoverable scene states}: robot actions change the scene such that the policy can no longer recover (e.g., severe object displacement, entanglement, or clutter accumulation).
\end{enumerate}
When an unrecoverable state occurs, we apply the minimum intervention needed to resume execution from the last recoverable state rather than resetting the task from scratch. This metric captures both safety interruptions and practical autonomy breakdowns under a realistic deployment protocol.

\subsubsection{Training Data}

Although Cortex~2.0 transfers efficiently to new tasks due to extensive pretraining on real-world deployment data, we collected targeted demonstrations for each benchmark task to enable fair comparison with baselines. Table~\ref{tab:data} summarizes data volumes.

\begin{table}[h]
\centering
\caption{Demonstration data collected per benchmark task.}
\label{tab:data}
\begin{tabular}{lcccc}
\toprule
\textbf{Task} & \textbf{Episodes} & \textbf{Hours} & \textbf{Arms} & \textbf{Robot} \\
\midrule
Pick-and-Place & 560    & 0.5  & 1 & single-arm \\
Sorting Items and Trash  & 8{,}700 & 21.0 & 2 & dual-arm \\
Sorting Screws           & 3{,}100 & 8.2  & 2 & dual-arm \\
Shoebox                  & 2{,}900 & 8.1  & 2 & dual-arm \\
\bottomrule
\end{tabular}
\end{table}

\subsubsection{Planning Budget}

\begin{figure}[ht]
  \centering
  \includegraphics[width=0.8\textwidth]{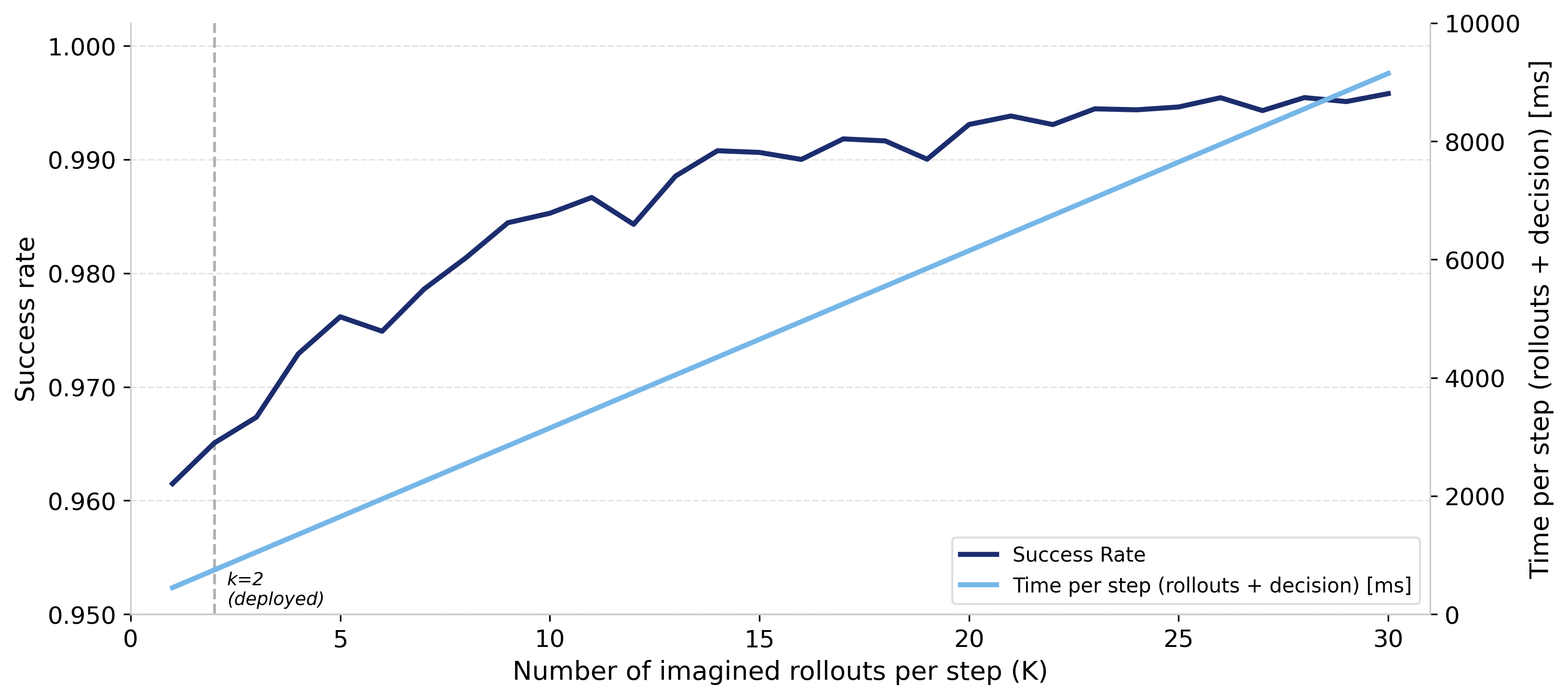}
  \caption{Cortex 2.0 Performance against Number of Rollouts $k$: With increasing number of rollouts $k$, the performance increases from 0.962 with 1 rollout to 0.996 at 30 rollouts. At the same time the time per step increases from ~310 ms at a single rollout to ~9200 ms for 30 rollouts.}
  \label{fig:tradeoff}
\end{figure}

A key design parameter in Cortex~2.0 is $k$, the number of imagined future rollouts sampled and scored by PRO before committing to an action. Success rate increases with $k$ while inference time per step increases linearly, capturing the central trade-off between foresight quality and computational cost (Figure~\ref{fig:tradeoff}). 

For all task evaluations below we fix a low-latency setting of $k{=}2$. The budget can be adjusted per task: higher $k$ for costly failure modes such as packing, where errors compound, and lower $k$ when recovery is cheap such as regrasping. Beyond $k$, rollout quality can be independently controlled via the number of denoising steps in the flow-matching world model, providing a second axis for the compute--quality trade-off.

\subsection{Task Definitions}

\begin{figure}[h]
    \centering
    \begin{tikzpicture}
        \node[inner sep=0] (img1) at (0,0) 
            {\includegraphics[width=0.28\linewidth]{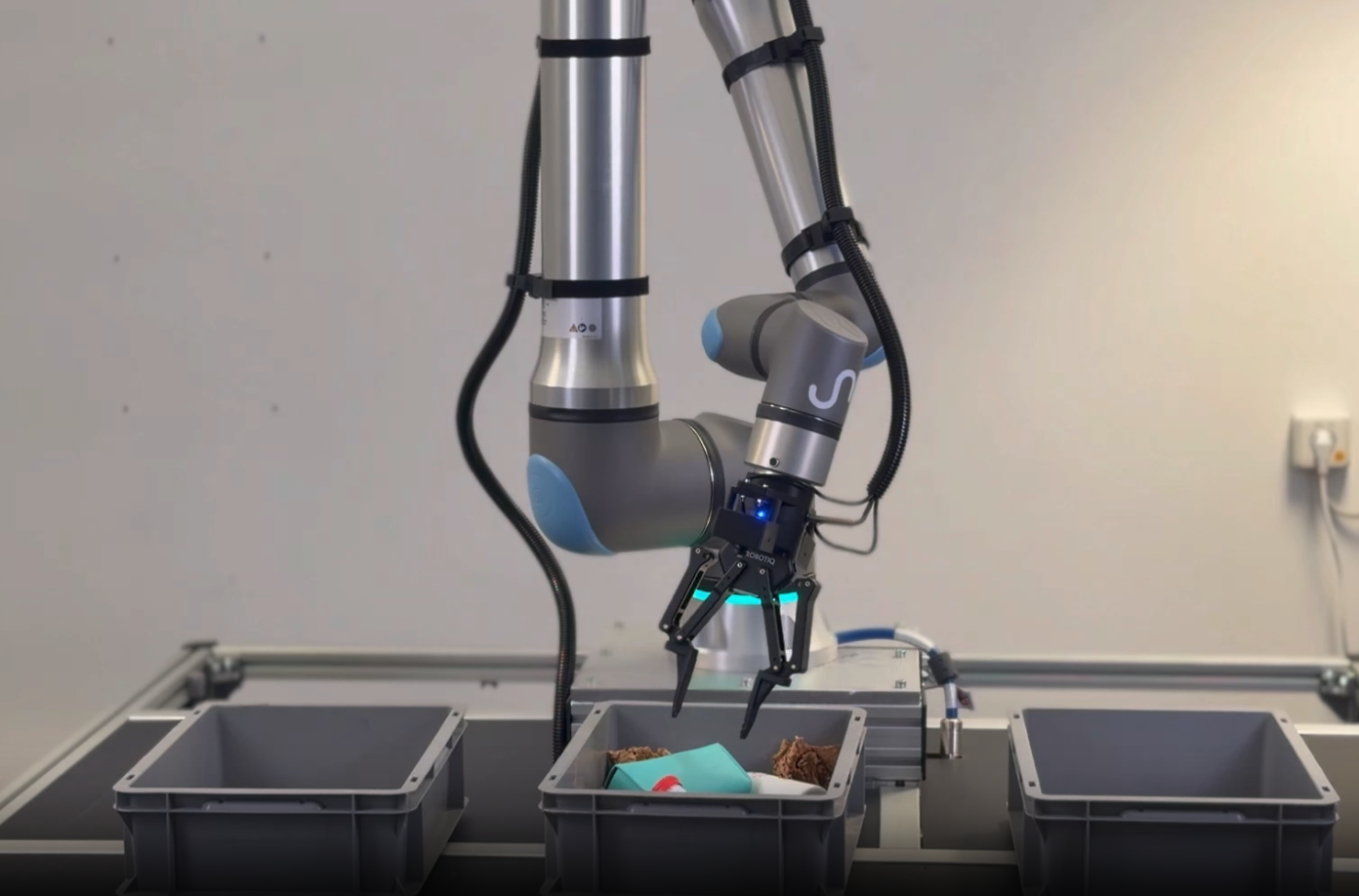}};
        \node[inner sep=0] (img2) at (5,0) 
            {\includegraphics[width=0.28\linewidth]{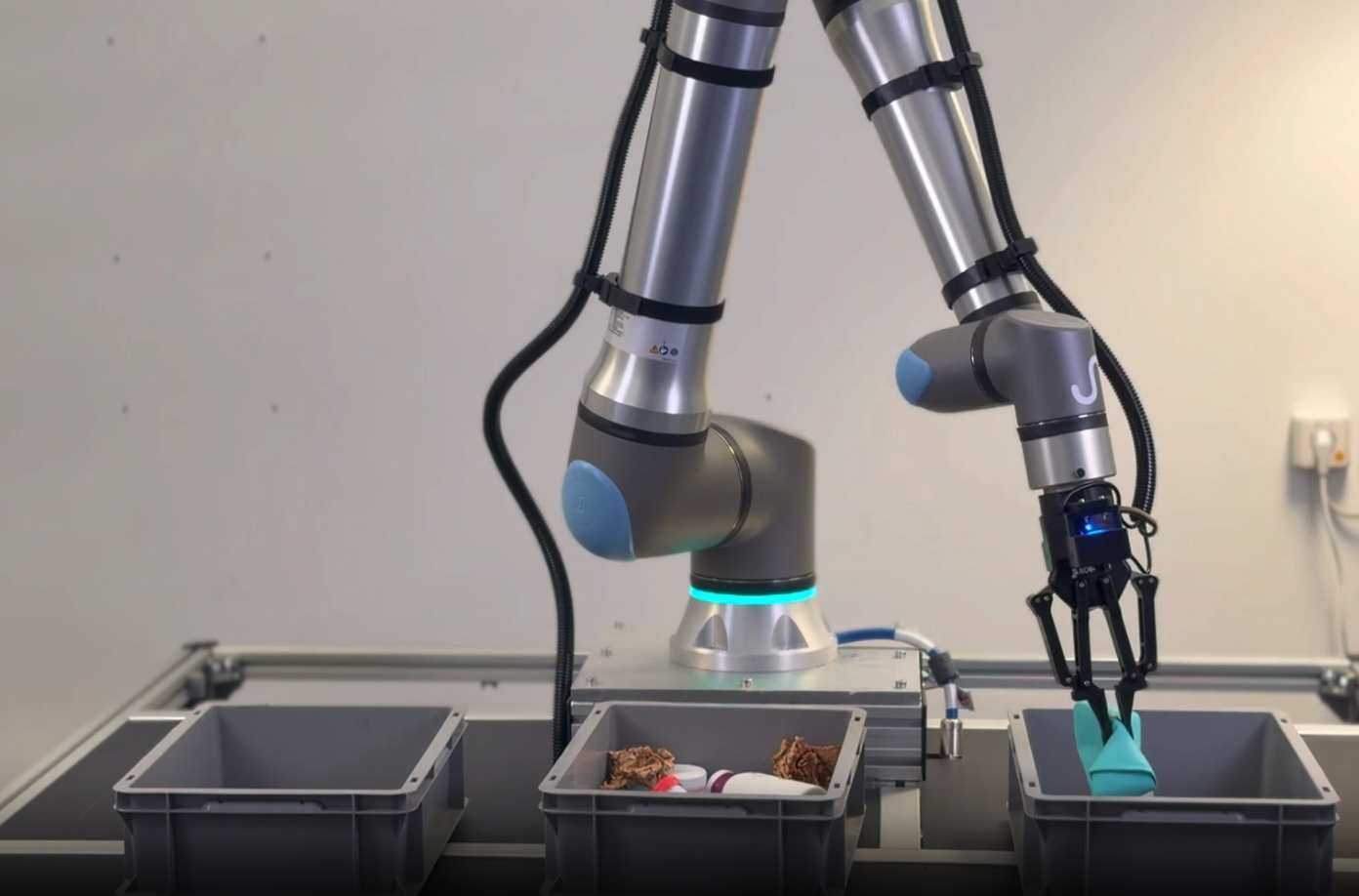}};
        \node[inner sep=0] (img3) at (10,0) 
            {\includegraphics[width=0.28\linewidth]{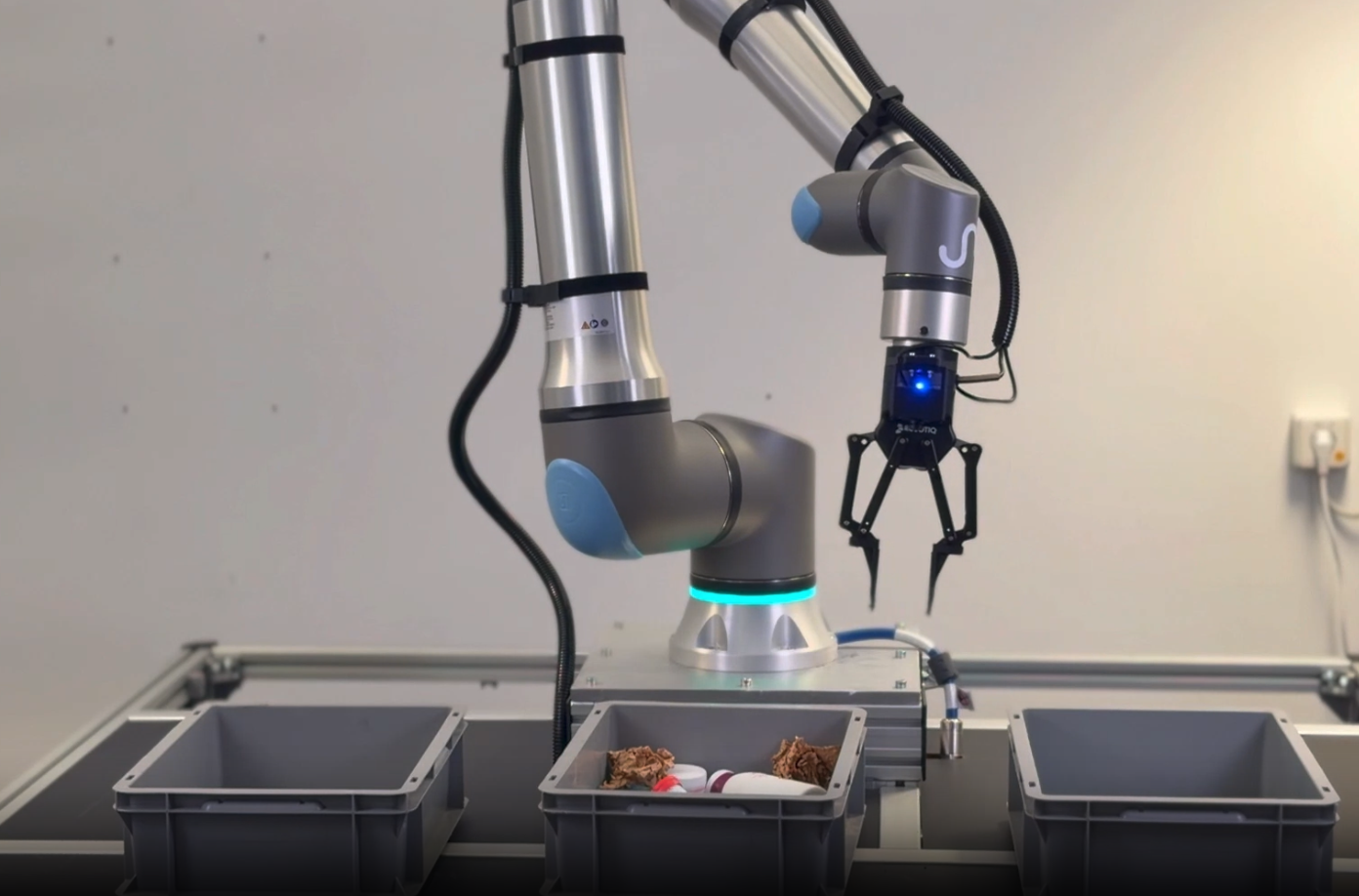}};
        
        \draw[->, thick] (img1.east) -- (img2.west);
        \draw[->, thick] (img2.east) -- (img3.west);
    \end{tikzpicture}
    \caption{Subtasks of Single-Arm Pick-and-Place. The robot detects best item to pick, transitions to the left bin and releases the object before moving to its home position.}
    \label{fig:three_images}
\end{figure}

\paragraph{Single-Arm Pick-and-Place.}
As illustrated in Figure~\ref{fig:three_images}, the robot grasps an item from a source bin and places it in a target bin. We fine-tune on 160 episodes. Trials receive 1.0 for success and fractional credit for partial completion; we report the mean over 16 trials. Although the simplest task in our benchmark suite, it still exposes characteristic failure modes of reactive policies: cluttered bins or difficult object poses lead to collisions, regrasp deadlocks, and releases that leave the item unrecoverable.

\begin{figure}[h]
    \centering
    \begin{tikzpicture}
        \node[inner sep=0] (img1) at (0,0) 
            {\includegraphics[width=0.28\linewidth]{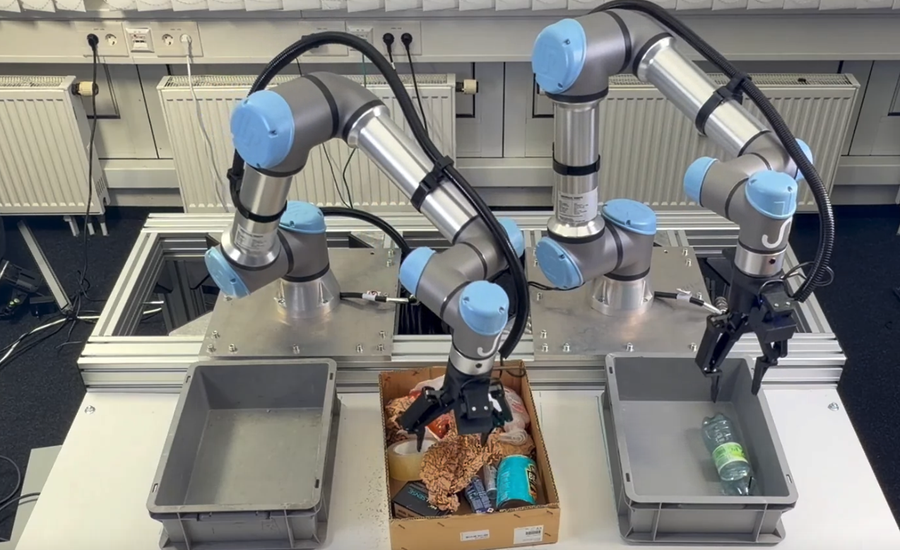}};
        \node[inner sep=0] (img2) at (5,0) 
            {\includegraphics[width=0.28\linewidth]{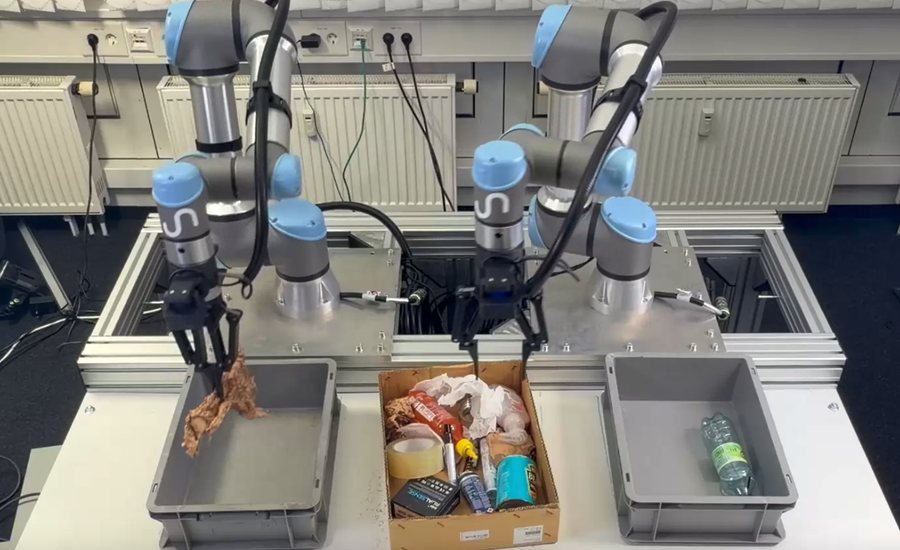}};
        \node[inner sep=0] (img3) at (10,0) 
            {\includegraphics[width=0.28\linewidth]{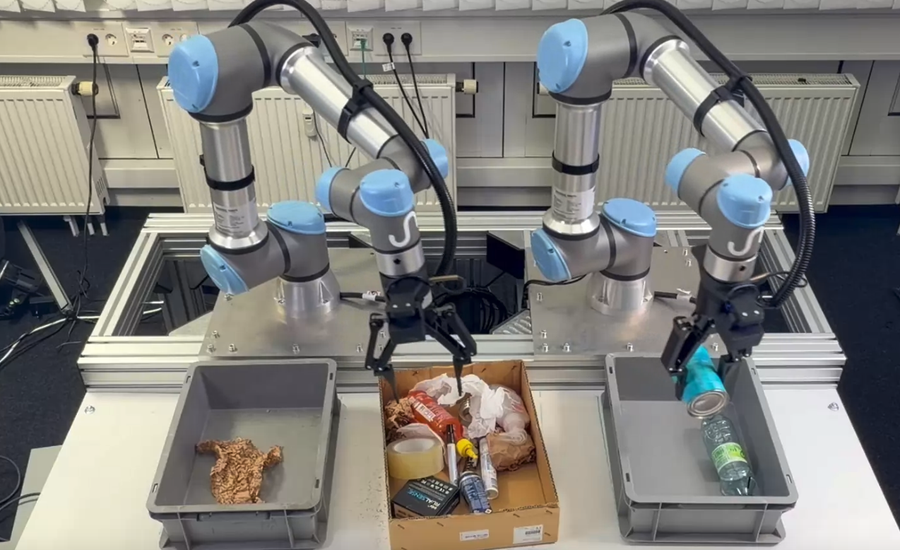}};
        
        \draw[->, thick] (img1.east) -- (img2.west);
        \draw[->, thick] (img2.east) -- (img3.west);
    \end{tikzpicture}
    \caption{Subtasks Sorting Items and Trash. The robot must detect which items are trash and which ones are good, then place trash into right bin and good items into left bin.}
    \label{fig:three_images1}
\end{figure}

\paragraph{Sorting Items and Trash.}
The dual-arm robot is presented with a cardboard box containing 10--15 randomly placed items and trash. The objective is to sort contents by placing trash into a left bin and items into a right bin (Figure~\ref{fig:three_images1}). This evaluates category discrimination in clutter, grasp planning, and reliable pick-and-place across diverse object types. A trial is considered successful only when all items are correctly sorted into their respective bins. Any operation requiring human intervention is counted as a failure. The cluttered source bin forces the robot to navigate tight spaces, where imprecise approach or extraction motions frequently result in collisions with the bin walls. Category mistakes such as placing trash into the items bin are common in the presence of visually ambiguous objects.

\begin{figure}[h]
    \centering
    \begin{tikzpicture}
        \node[inner sep=0] (img1) at (0,0) 
            {\includegraphics[width=0.28\linewidth]{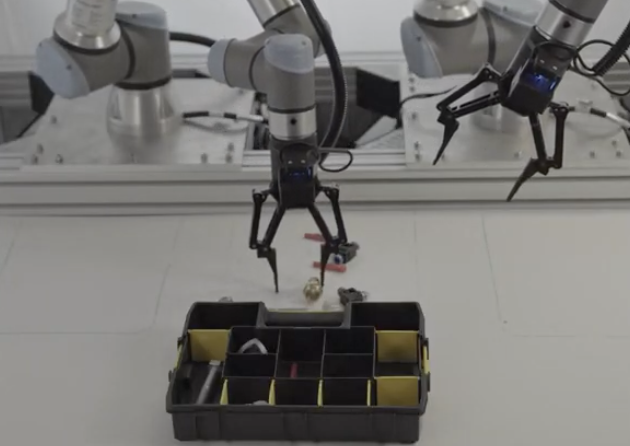}};
        \node[inner sep=0] (img2) at (5,0) 
            {\includegraphics[width=0.28\linewidth]{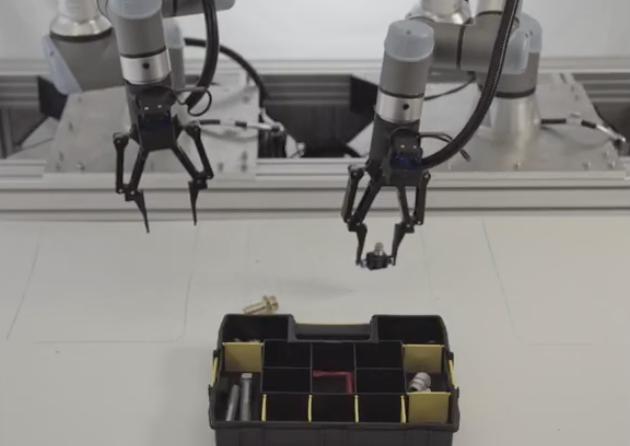}};
        \node[inner sep=0] (img3) at (10,0) 
            {\includegraphics[width=0.28\linewidth]{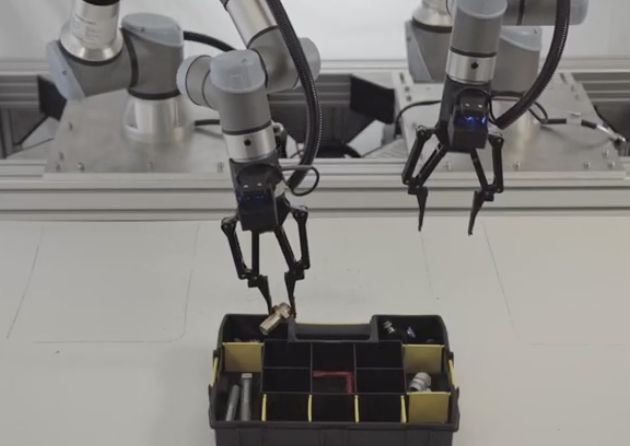}};
        
        \draw[->, thick] (img1.east) -- (img2.west);
        \draw[->, thick] (img2.east) -- (img3.west);
    \end{tikzpicture}
    \caption{Subtasks of Sorting Screws. All screws are sorted into the according compartments.}
    \label{fig:three_images2}
\end{figure}

\paragraph{Sorting Screws.}
Fine-grained manipulation of small metal screws and tooling scattered on a tabletop is shown in Figure~\ref{fig:three_images2}. The robot picks each screw and deposits it into the correct compartment of a multi-section toolbox. The challenge lies in precisely localizing and grasping small, reflective objects that may roll or shift during manipulation under challenging lighting conditions. A trial is considered successful only when all screws are correctly placed into their respective compartments. This task is particularly hard for visuomotor policies: screws are small, reflective, and can be occluded by one another. Sub-millimeter misalignments during approach shift the screw rather than grasping it. Human interventions are required when a dropped screw rolls outside the reachable workspace and must be physically returned, when the policy repeatedly attempts to grasp an unreachable or occluded screw without recovering, and when a screw is deposited into the wrong compartment and must be manually corrected.

\begin{figure}[h]
    \centering
    \begin{tikzpicture}
        \node[inner sep=0] (img1) at (0,0) 
            {\includegraphics[width=0.28\linewidth]{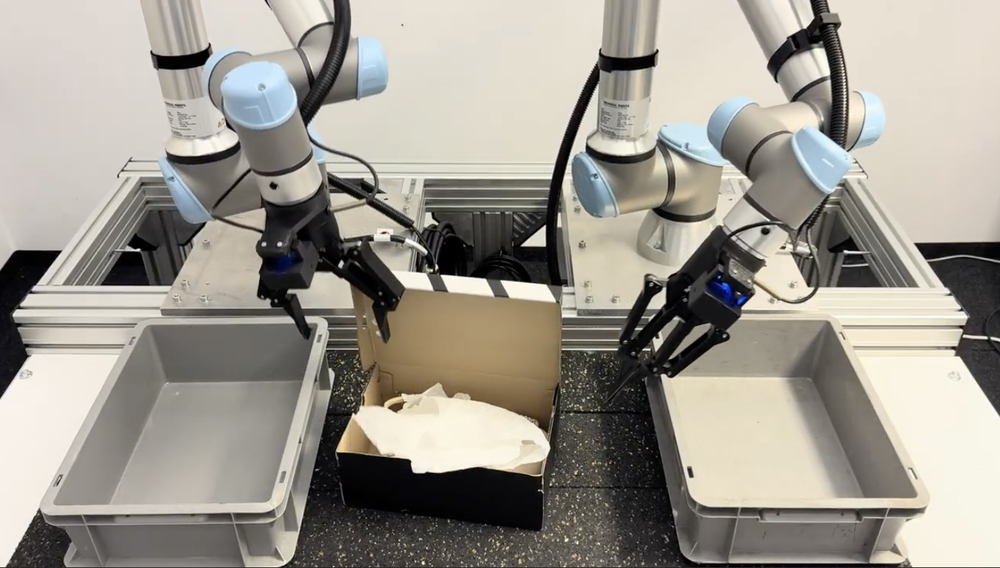}};
        \node[inner sep=0] (img2) at (5,0) 
            {\includegraphics[width=0.28\linewidth]{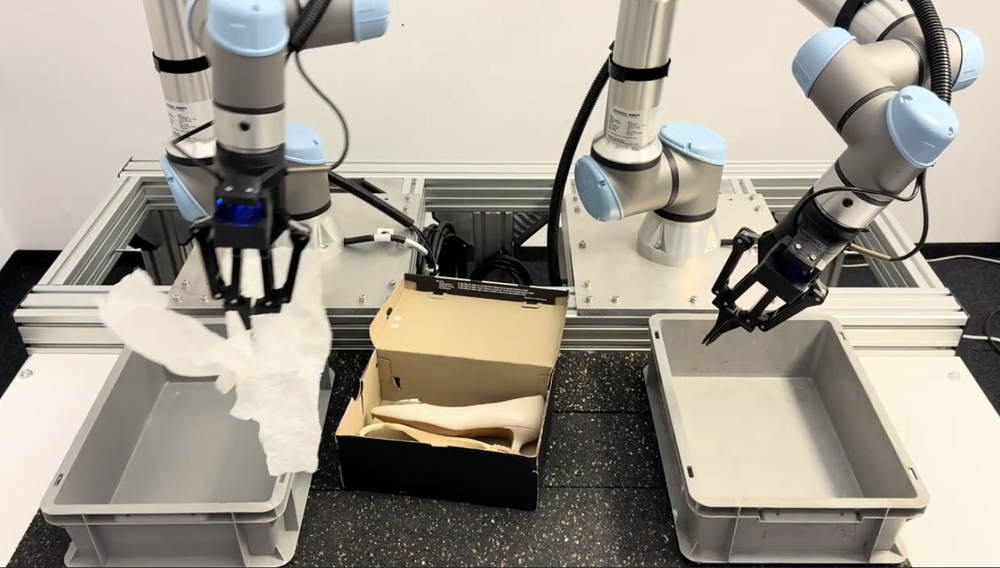}};
        \node[inner sep=0] (img3) at (10,0) 
            {\includegraphics[width=0.28\linewidth]{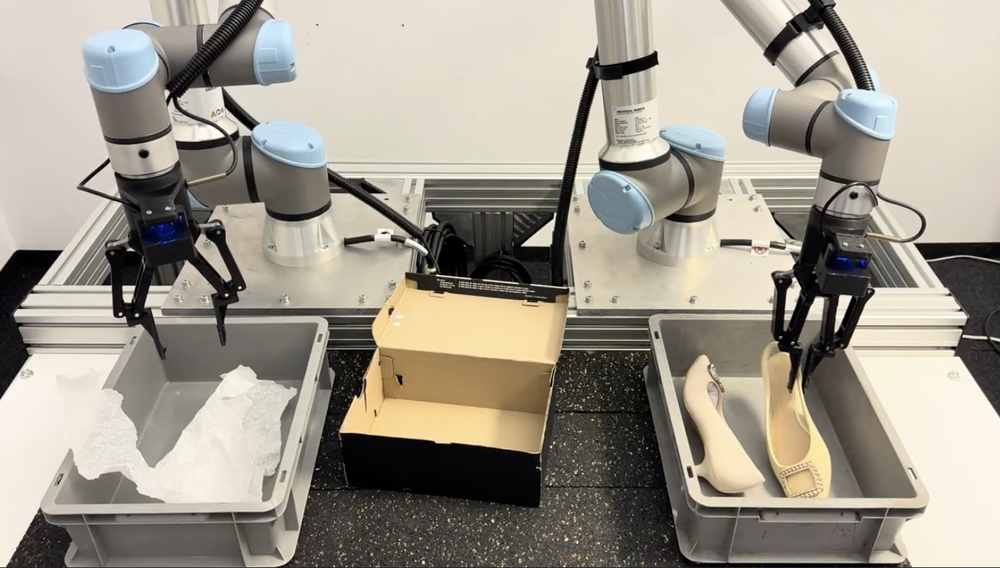}};
        
        \draw[->, thick] (img1.east) -- (img2.west);
        \draw[->, thick] (img2.east) -- (img3.west);
    \end{tikzpicture}
    \caption{Subtasks of Shoebox Unpacking. The shoebox is opened, packaging paper is removed and both shoes are placed into the right bin.}
    \label{fig:three_images3}
\end{figure}

\paragraph{Shoebox Unpacking.}
The robot executes a four-step sequence: (1)~open a closed shoebox lid; (2)~remove packing paper and place it in the left bin; (3)~extract one shoe and place it in the right bin; (4)~extract the remaining shoe and place it in the right bin (Figure~\ref{fig:three_images3}). This tests long-horizon execution, deformable object handling, and articulated container manipulation. Success requires completion of all steps; partial completion is not counted. The robot must first identify the correct side of the shoebox to open; pressing or pulling on the wrong face can punch a hole through the cardboard. Once open, every subtask reshapes the scene: as the lid is opened or the paper is removed, new information about the next step emerges. Interventions are triggered when the robot fails to open the correct side, when a shoe slips from an unstable grasp and falls outside the bin, or when the second shoe is never located and the policy stops making progress.

\subsection{Results}

\subsubsection{Experiment 1: Single-Arm Pick-and-Place}

The pick-and-place task evaluates precision manipulation and low-data adaptation across 16 trials, with item position and orientation randomized between trials. Scoring is continuous: trials receive 1.0 for full success and fractional credit for partial completion.

Cortex~2.0 achieves the highest mean score across all methods, demonstrating strong precision under the low-data regime of only 160 fine-tuning episodes. $\pi_{0.5}$ achieves the second-highest score but exhibits greater variance across trials, reflecting inconsistent approach trajectories when adapting from its pretraining distribution to the single-arm setting. Diffusion Policy attains a moderate mean score; its primary failure mode are placement misalignment at the final stage and premature object drops during approach. RDT-2 achieves the lowest mean score, with frequent grasp failures.

\begin{table}[h]
\centering
\caption{Single-arm pick-and-place results. Mean score over 16 trials; 1.0 = full success, fractional credit for partial completion.}
\label{tab:pickplace}
\begin{tabular}{lccc}
\toprule
\textbf{Model} & \textbf{Success Rate} & \textbf{Avg. Completion Time} & \textbf{Human Interventions} \\
\midrule
Cortex 2.0 (ours)                        & \textbf{0.98} & \textbf{20s} & \textbf{0}  \\
$\pi_{0.5}$~\cite{pi05}                  & 0.7          & 49s           & 2           \\
Diffusion Policy~\cite{diffusionpolicy} & 0.56          & 53s           & 4           \\
RDT-2~\cite{liu2026rdt2}                  & 0.4          & 63s           & 7           \\
\bottomrule
\end{tabular}
\end{table}

\subsubsection{Experiment 2: Sorting Items and Trash}

The sorting task evaluates repeated pick-and-place in a cluttered environment across 10~rollouts per policy. Each rollout begins with 10--15 objects randomly placed in the source box, with object types, quantities, positions, and orientations randomized between rollouts. 

Cortex~2.0 achieves the highest per-operation success rate and the shortest average task duration, completing all rollouts without any human intervention. In contrast, all baseline policies require human intervention in every rollout to complete the task. $\pi_{0.5}$ achieves higher success than the remaining baselines but fails to complete the full task within the 15-minute execution limit in all runs; its dominant failure mode is repeated local replanning around failed grasp attempts. Diffusion Policy attains moderate success but with greater instability driven by grasp failures and object drops. RDT-2 achieves significantly lower success rates than all other methods. Both Diffusion Policy and RDT-2 also reach the 15-minute execution limit in all runs without completing the full task.

Cortex~2.0 is qualitatively different from all baselines in this experiment: it reliably completes the task autonomously, whereas all baseline methods depend on repeated human intervention to finish execution.

\begin{table}[h]
\centering
\caption{Sorting items and trash results. Success reported per individual sorting operation.}
\label{tab:sorting}
\begin{tabular}{lccc}
\toprule
\textbf{Model} & \textbf{Per-Op.\ Success} & \textbf{Task Completion} & \textbf{Human Interventions} \\
\midrule
Cortex 2.0 (ours)                        & \textbf{0.95} & \textbf{700s} & \textbf{0}  \\
$\pi_{0.5}$~\cite{pi05}                  & 0.61 & —$^{*}$  & 53 \\
Diffusion Policy~\cite{diffusionpolicy} & 0.47 & —$^{*}$ & 59 \\
RDT-2~\cite{liu2026rdt2}                  & 0.18  & —$^{*}$  & 95 \\
\bottomrule
\end{tabular}
\\[0.5em]
\small $^{*}$ Task not completed within the execution limit.
\end{table}

\subsubsection{Experiment 3: Sorting Screws}

The screw sorting task evaluates fine-grained manipulation across 10~rollouts per policy, with screw positions and orientations randomized between rollouts. Performance is reported as per-operation success rate aggregated across all screws and all rollouts. 

Cortex~2.0 substantially outperforms all baseline methods, achieving near-perfect per-operation success while completing the task in the shortest average time and without entering any unrecoverable states. Among the baselines, this task exhibits the largest performance gap relative to Cortex~2.0. $\pi_{0.5}$ attains moderate success and remains the strongest baseline but exhibits limited precision on small objects of varying shapes. Diffusion Policy achieves lower success rates than $\pi_{0.5}$ but demonstrates comparatively greater stability. RDT-2 fails almost entirely, with zero successful placements and multiple unrecoverable states per rollout.

These results highlight the direct benefit of PRO-based lookahead in precision-critical scenarios: by filtering rollouts that pass through latent states associated with unstable contact, the system avoids the small errors that would otherwise shift object pose and compound task difficulty in subsequent steps.

\begin{table}[h]
\centering
\caption{Screw sorting results. Success reported per individual screw placement.}
\label{tab:screws}
\begin{tabular}{lccc}
\toprule
\textbf{Model} & \textbf{Per-Op.\ Success} & \textbf{Avg.\ Completion Time} & \textbf{Human Interventions} \\
\midrule
Cortex 2.0 (ours)                        & \textbf{0.98} & \textbf{180\,s} & \textbf{0}  \\
$\pi_{0.5}$~\cite{pi05}                  & 0.4 & —$^{*}$ & 24 \\
Diffusion Policy~\cite{diffusionpolicy} & 0.2 & —$^{*}$ & 16 \\
RDT-2~\cite{liu2026rdt2}                  & 0.0  & —$^{*}$  & 50 \\
\bottomrule
\end{tabular}
\\[0.5em]
\small $^{*}$ Task not completed within the execution limit.
\end{table}

\subsubsection{Experiment 4: Shoebox Unpacking}

The shoebox task evaluates multi-step manipulation capabilities across 10~rollouts per policy, with box orientation, paper configuration, and shoe placement randomized between rollouts. 

Cortex~2.0 achieves the highest holistic success rate while completing the task significantly faster than all baseline methods, and does so without requiring human intervention across any rollout. This indicates strong temporal consistency and the ability to maintain task context across the full four-step sequence. Notably, this benchmark is trained with significantly fewer demonstrations than the sorting tasks, emphasizing Cortex~2.0's ability to transfer and scale to complex task structure in a data-limited setting.

$\pi_{0.5}$ achieves relatively high success at the subtask level but fails to complete the full task more frequently than Cortex~2.0, requires substantially longer execution times, and frequently enters unrecoverable states in later stages. It often progresses through the early steps but struggles to adapt to scene changes introduced by prior actions. Diffusion Policy exhibits significantly lower success rates, failing primarily during shoe extraction due to unstable grasping and limited recovery from failed picks. RDT-2 never completes the full task sequence and frequently enters unrecoverable states, preventing meaningful progress beyond the early stages.

\begin{table}[h]
\centering
\caption{Shoebox unpacking results. Full task success requires completion of all four sequential steps.}
\label{tab:shoebox}
\begin{tabular}{lccc}
\toprule
\textbf{Model} & \textbf{Success Rate} & \textbf{Avg.\ Completion Time} & \textbf{Human Interventions} \\
\midrule
Cortex 2.0 (ours)                        & \textbf{0.96}  & \textbf{58s} & \textbf{0}  \\
$\pi_{0.5}$~\cite{pi05}                  & 0.6          & 103s & 5 \\
Diffusion Policy~\cite{diffusionpolicy} & 0.12              & 52s               & 9 \\
RDT-2~\cite{liu2026rdt2}                  & 0.0             & 62s             & 10 \\
\bottomrule
\end{tabular}
\end{table}

\subsection{Toward In-Context Learning}
Making the policy video-aware and training a world model jointly with policy learning constitutes a step toward in-context learning for robotics. Given a short sequence of demonstrations in the form of a video, the robot should be able to execute analogous steps without retraining. Today’s large language models exhibit in-context learning capabilities across many applications: agents can be conditioned to execute tasks through language alone. We are working toward analogous capabilities for robots, where visual demonstrations serve as task specifications.
Formally, this corresponds to a policy $\pi_\theta(a \mid o, \tau^{\text{demo}})$ conditioned on a video demonstration $\tau^{\text{demo}}$ alongside the current observation $o$. The world model provides a natural substrate for this: by learning rich visual dynamics, it develops representations that support analogical reasoning over demonstration sequences, unlocking a new dimension of generalization for physical AI.
\subsection{Summary and Discussion}

Across all four benchmarks, Cortex~2.0 consistently achieves higher success rates and shorter execution times than all evaluated baseline policies, and is the only method to complete tasks without human intervention.

$\pi_{0.5}$ generally attains the strongest performance among the baselines at the subtask level but requires substantially longer execution times and frequently fails to complete tasks end-to-end, particularly in long-horizon settings. Diffusion Policy exhibits lower overall success, while RDT-2 consistently fails to complete complex tasks. All three baseline policies enter unrecoverable states more frequently than Cortex~2.0, reducing overall reliability and increasing dependence on human intervention.
\begin{figure}[h]
    \centering
    \begin{subfigure}[b]{0.32\textwidth}
        \includegraphics[width=\linewidth]{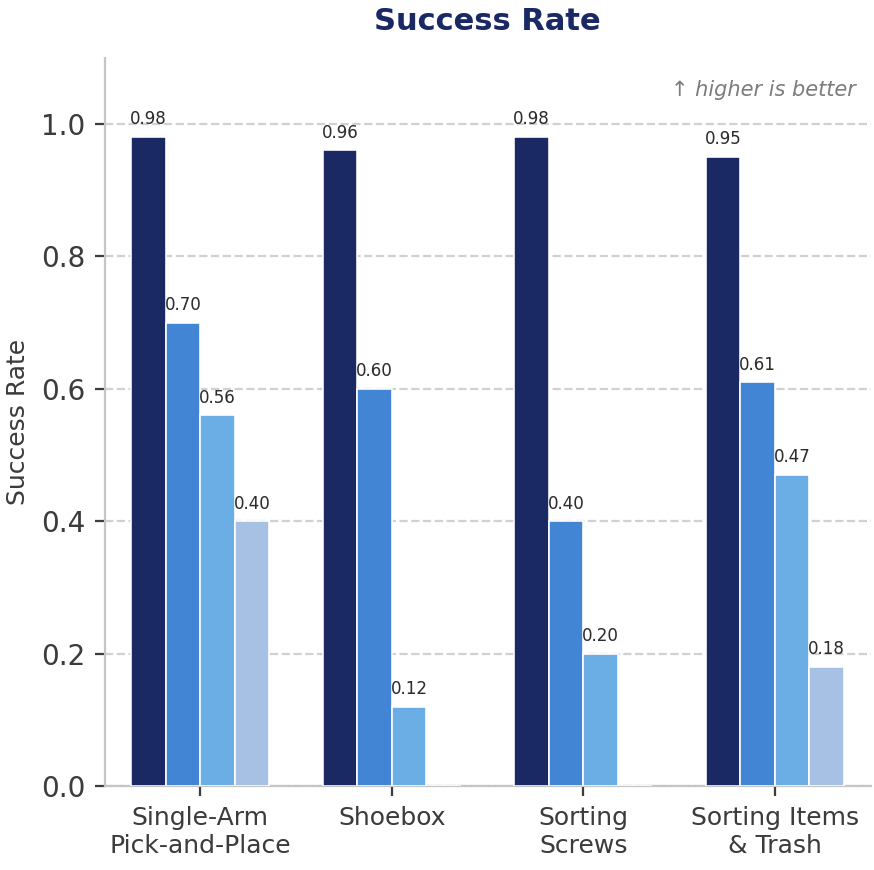}
    \end{subfigure}
    \hfill
    \begin{subfigure}[b]{0.32\textwidth}
        \includegraphics[width=\linewidth]{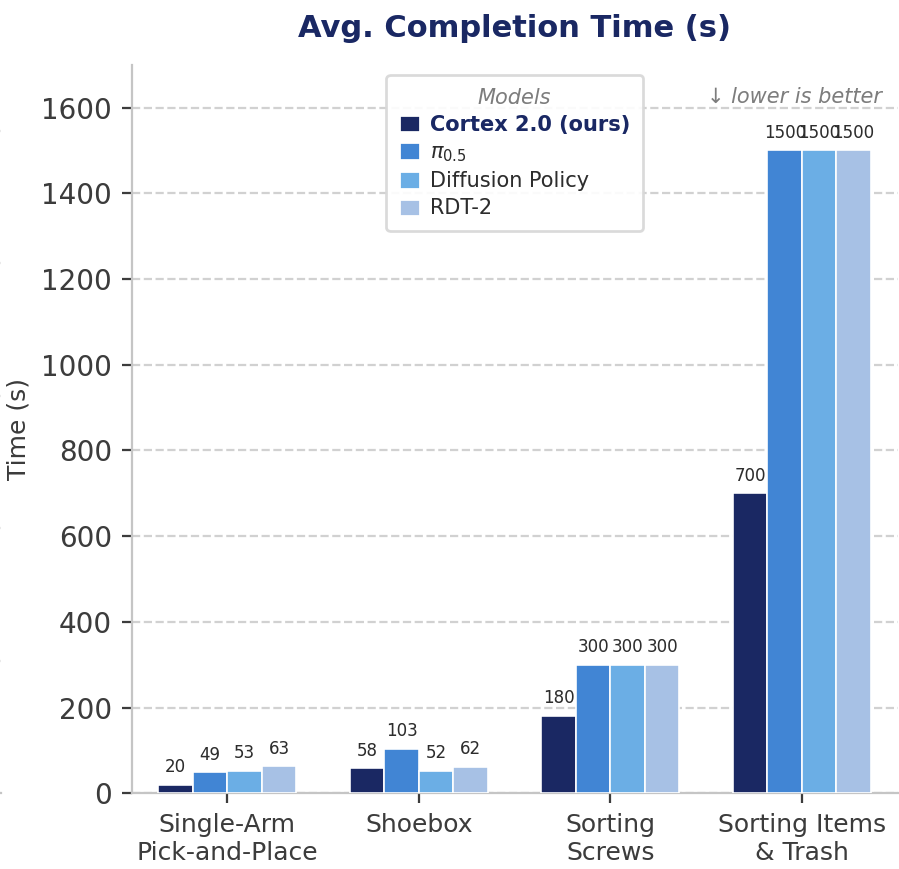}
    \end{subfigure}
    \hfill
    \begin{subfigure}[b]{0.32\textwidth}
        \includegraphics[width=\linewidth]{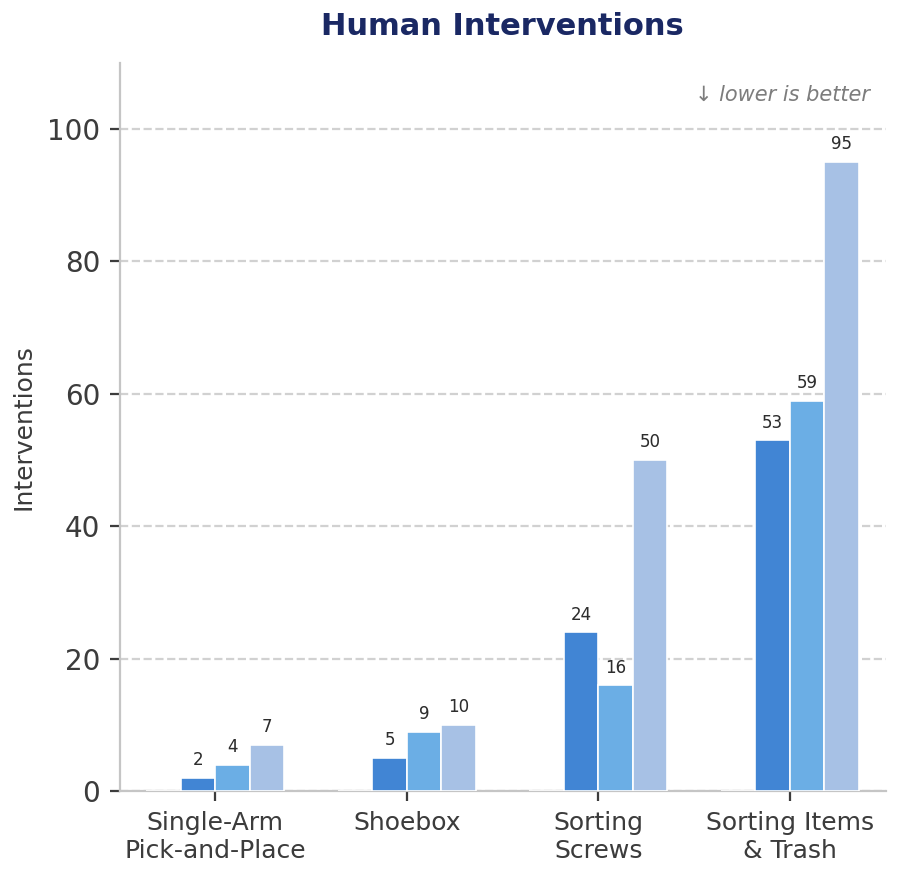}
    \end{subfigure}
    \caption{Performance comparison across all four benchmark tasks.
    \textbf{Left:} success rate (higher is better).
    \textbf{Centre:} average completion time in seconds (lower is better;
    baselines are capped at the 1{,}500\,s execution limit for the item sorting tasks and at 300\,s for the screw sorting task).
    \textbf{Right:} total human interventions per task (lower is better).}
    \label{fig:benchmark_summary}
\end{figure}

The results support two key claims of Cortex~2.0. First, pretraining on real-world deployment data enables strong generalization: with limited fine-tuning data, Cortex~2.0 achieves success rates beyond 90\% on all tasks, enabling a strong baseline model at deployment that reaches 99\% after continued operation on in-domain data. Second, world-model-based planning provides qualitative robustness benefits beyond success rate: by filtering bad branches before execution, the system avoids the characteristic failure mode of reactive baselines, in which a missed grasp triggers repeated retries and eventually escalates into deadlock. This directly translates into zero human interventions across all benchmarks, a metric that more accurately reflects the industrial operational cost than the success rate alone.
\section{Conclusion}

Deploying generalist robot policies in real production environments remains a difficult challenge: reactive systems fail under long-horizon task sequences, compound errors are costly to recover from, and the diversity of objects, layouts, and embodiments cannot be fully covered by any finite dataset. Our work with Cortex 2.0 shows that vision–language–action models can overcome many of these barriers when augmented with physical foresight.
The central contribution of Cortex 2.0 is the integration of a world model into the manipulation policy loop, shifting from try-and-see control to plan-and-try. By generating candidate futures in visual latent space and scoring them via PRO before any action is executed, the system filters out bad branches before they become unrecoverable states. Across warehouse pick-and-place, item sorting, and shoebox handling evaluations, Cortex 2.0 consistently outperformed all baselines while requiring zero human interventions. The planning budget $k$ provides a practical lever to trade foresight quality against latency, enabling the same system to allocate more computation to high-stakes decisions such as packing, and less to cheap-recovery situations such as regrasping.

Training in visual latent space proved critical for two reasons. First, it enables cross-embodiment transfer: the same PRO scoring function and world model operate across single-arm, dual-arm, and other robot platforms without modification, because visual representations encode transferable physical regularities that are independent of specific kinematics. Second, it makes the data strategy scalable: deployment cameras provide a natural multiplier on training signal, and the continuous feedback loop from live operations ensures the world model learns from the full complexity of real-world conditions rather than simplified lab setups.

We are continuously expanding the deployment database with new operational data from our robot fleet, expanding coverage to new task families, object categories, and embodiments, and rolling out the policy to new industrial partners and workflows. Each deployment adds training signal that feeds back into the world model and PRO, compounding improvements in planning quality over time. The gap between a research system and a production system is not closed at training time but closed through expanded deployment, iteration, and scale.

\textbf{Future work.} 
We plan to scale world model training substantially, both in compute and 
data. Allocating more training time and leveraging a larger portion of 
our collected deployment data can improve the fidelity of predicted rollouts.
Since the VLA policy is conditioned on the selected future latent, sharper predictions provide a more informative signal to the policy and directly improve the generated action chunk.
Current evaluations cover a subset of embodiments and task families, 
and the planning horizon $H_{wm}$ and budget $k$ are fixed per task. We therefore 
plan to strengthen online adaptation and uncertainty-aware dynamic budget 
allocation in PRO, tighten the coupling between video tokenization and 
control for longer-horizon foresight, and validate in-context learning from 
video demonstrations on unseen task families at test time.
We believe Cortex 2.0 is a step toward dependable, general-purpose robot 
intelligence that plans before it acts, adapts continuously from deployment, 
and scales robustly to the messy, continuously changing conditions of real 
production environments.

\bibliographystyle{unsrtnat}
\bibliography{refs}






\end{document}